\documentclass[11pt,a4paper]{article}
\usepackage{times,latexsym}
\usepackage{url}
\usepackage[T1]{fontenc}
\usepackage{graphicx}
\usepackage{multirow}
\usepackage[british]{babel}
\usepackage{amsmath}

\usepackage[acceptedWithA]{tacl2018v2}
\usepackage[]{tacl2018v2}

\usepackage{xspace,mfirstuc,tabulary}

\newif\iftaclinstructions
\taclinstructionsfalse 
\iftaclinstructions
\renewcommand{\confidential}{}
\renewcommand{\anonsubtext}{(No author info supplied here, for consistency with
TACL-submission anonymization requirements)}
\newcommand{\instr}
\fi

\iftaclpubformat 

\else

\fi

\title{Characterizing English Variation across \\ Social Media Communities with BERT}

\author{
 Li Lucy \and David Bamman\\
 University of California, Berkeley \\
  {\sf \{lucy3\_li, dbamman\}@berkeley.edu} \\
}

\date{}

\begin{document}
\maketitle
\begin{abstract}
Much previous work characterizing language variation across Internet social groups has focused on the \textit{types} of words used by these groups. We extend this type of study by employing BERT to characterize variation in the \textit{senses} of words as well, analyzing two months of English comments in 474 Reddit communities. The specificity of different sense clusters to a community, combined with the specificity of a community's unique word types, is used to identify cases where a social group's language deviates from the norm. We validate our metrics using user-created glossaries and draw on sociolinguistic theories to connect language variation with trends in community behavior. We find that communities with highly distinctive language are medium-sized, and their loyal and highly engaged users interact in dense networks.
\end{abstract}

\begin{figure}[ht]
	\centering{\includegraphics[width=0.8\linewidth]{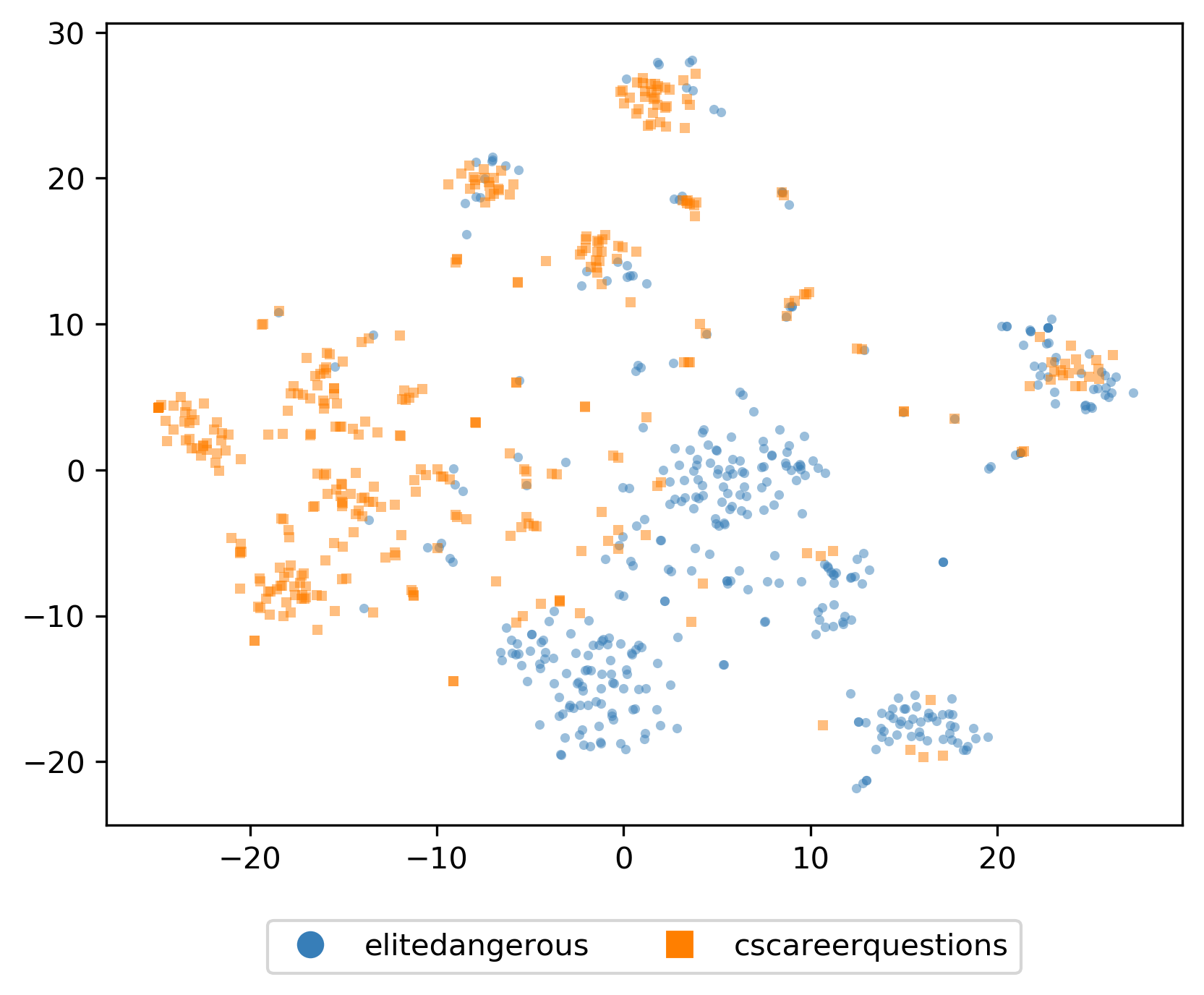}}
	\centering{\includegraphics[width=0.9\linewidth]{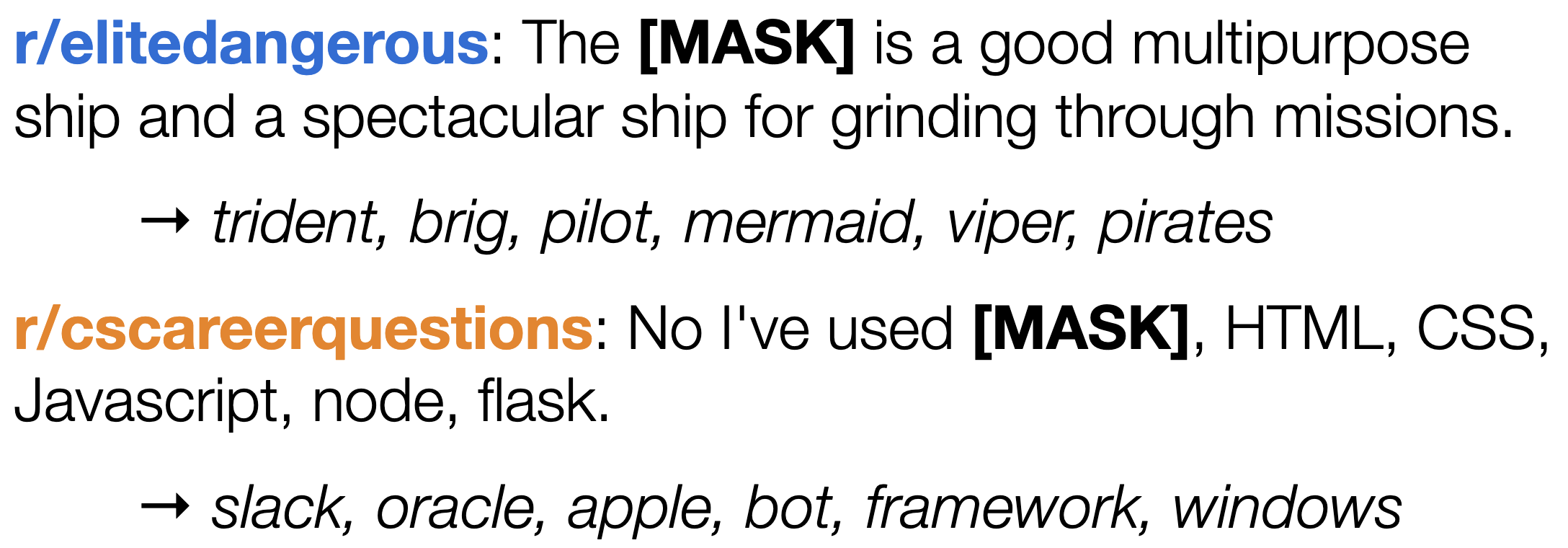}}
	\caption{Different online communities may systematically use the same word to mean different things. Each marker on the t-SNE plot is a BERT embedding of \textit{python}, case insensitive, in r/cscareerquestions (where it refers to the programming language) and r/elitedangerous (where it refers to a type of spacecraft). BERT also predicts different substitutes when \textit{python} is masked out in these communities' comments.}
	\label{fig:fig_1}
\end{figure}

\section{Introduction}

Internet language is often popularly characterized as a messy variant of ``standard'' language \cite{mashable, unbabel}. However, work in sociolinguistics has demonstrated that online language is not homogeneous \cite{JOSL:JOSL287,nguyen2016computational, eisenstein-2013-bad}. Instead, it expresses immense amounts of variation, often driven by social variables. Online language contains lexical innovations, such as orthographic variants, but also repurposes words with new meanings \cite{pei2019slang, stewart2017anorexia}. There has been much attention on \textit{which} words are used across these social groups, including work examining the frequency of types \cite{zhang2017community, danescu2013no}. However, there is also increasing interest in \textit{how} words are used in these online communities as well, including variation in meaning \cite{yang2017overcoming, del2017semantic}. For example, a word such as \textit{python} in Figure \ref{fig:fig_1} has different usages depending on the community in which it is used. Our work examines both lexical and semantic variation, and operationalizes the study of the latter using BERT \cite{devlin2019bert}. 

Social media language is an especially interesting domain for studying lexical semantics because users' word use is far more dynamic and varied than is typically captured in standard sense inventories like WordNet. Online communities that sustain linguistic norms have been characterized as virtual communities of practice \cite{eckert1992think, del2017semantic, nguyen2011language}. Users may develop wiki pages, or guides, for their communities that outline specific jargon and rules. However, some communities exhibit more language variation than others. One central goal in sociolinguistics is to investigate what social factors lead to variation, and how they relate to the growth and maintenance of sociolects, registers, and styles. To enable our ability to answer these types of questions from a computational perspective, we must first develop metrics for measuring variation. 

Our work quantifies how much the language of an online community deviates from the norm and identifies communities that contain unique language varieties. We define community-specific language in two ways, one based on word choice variation, and another based on meaning variation using BERT. Words used with community-specific senses match words that appear in glossaries created by users for their communities. Finally, we test several hypotheses about user-based attributes of online English varieties drawn from sociolinguistics literature, showing that communities with more distinctive language tend to be medium-sized and have more loyal and active users in dense interaction networks. We release our code, our dataset of glossaries for 57 Reddit communities, and additional information about all communities in our study at \url{https://github.com/lucy3/ingroup_lang}.

\section{Related Work}

The sheer number of conversations on social media platforms allow for large-scale studies that were previously impractical using traditional sociolinguistics methods such as ethnographic interviews and surveys. Earlier work on computer-mediated communication identified the presence and growth of group norms in online settings \cite{postmes2000formation}, and how new and veteran community members adapt to a community's changing linguistic landscape \cite{nguyen2011language, danescu2013no}. 

Much work in computational sociolinguistics has focused on lexical variation \cite{nguyen2016computational}. Online language contains an abundance of ``nonstandard'' words, and these dynamic trends rise and decline based on social and linguistic factors \cite{rotabi2016status, altmann2011niche, stewart2018making, del-tredici-fernandez-2018-road, eisenstein2014diffusion}. Online communities' linguistic norms and differences are often defined by \textit{which} words are used. For example, \citet{zhang2017community} quantify the distinctiveness of a Reddit community's identity by the average specificity of its words and utterances. They define specificity as the PMI of a word in a community relative to the entire set of communities, and find that distinctive communities are more likely to retain users. To identify community-specific language, we extend \citet{zhang2017community}'s approach to incorporate semantic variation, mirroring the sense-versus-type dichotomy of language variation put forth by previous work on slang detection \cite{dhuliawala2016slangnet,pei2019slang}. 

There has been less previous work on cross-community semantic variation. \citet{yang2017overcoming} use social networks to address sentiment variation across Twitter users, accounting for cases such as \textit{sick} being positive in \textit{this sick beat} or negative in \textit{I feel tired and sick}. \citet{del2017semantic} adapt the model of \citet{bamman2014distributed} for learning dialect-aware word vectors to Reddit communities discussing programming and football. They find that sub-communities for each topic share meaning conventions, but also develop their own. A line of future work suggested by \citet{del2017semantic} is extending studies on semantic variation to a larger set of communities, which our present work aims to achieve. 

The strength of BERT to capture word senses presents a new opportunity to measure semantic variation in online communities of practice. BERT embeddings have been shown to capture word meaning \cite{devlin2019bert}, and different senses tend to be segregated into different regions of BERT's embedding space \cite{wiedemann-etal-2019-does}. Clustering these embeddings can reveal sense variation and change, where distinct senses are often represented as cluster centroids \cite{hu2019diachronic,giulianelli-etal-2020-analysing}. For example, \citet{NIPS2019_9065} use a nearest-neighbor classifier for word sense disambiguation, where word embeddings are assigned to the nearest centroid representing a word sense. Using BERT-base, they achieve an F1 of 71.1 on SemCor \cite{miller1993semantic}, beating the state of the art at that time. Part of our work examines how well the default behavior of contextualized embeddings, as depicted in Figure \ref{fig:fig_1}, can be used for identifying niche meanings in the domain of Internet discussions. 

As online language may contain semantic innovations, our domain necessitates word sense induction (WSI) rather than disambiguation. We evaluate approaches for measuring usage or sense variation on two common WSI benchmarks, \mbox{SemEval} 2013 Task 13 \cite{jurgens-klapaftis-2013-semeval} and \mbox{SemEval} 2010 Task 14 \cite{manandhar-klapaftis-2009-semeval}, which provide evaluation metrics for unsupervised sense groupings of different occurrences of words. The current state of the art in WSI clusters representations consisting of substitutes, such as those shown in Figure~\ref{fig:fig_1}, predicted by BERT for masked target words \cite{amrami2018word,amrami2019towards}. We also adapt this method on our Reddit dataset to detect semantic variation. 

\begin{table*}[t]
\centering
\tiny
\resizebox{\textwidth}{!}{%
\begin{tabular}{l | c>{\raggedright}p{5cm}cc}
\hline \textbf{subreddit} & \textbf{word} & \textbf{definition} & \textbf{count} & \textbf{type NPMI}   \\ \hline
\multirow{ 4}{*}{r/justnomil} & fdh & ``future damn husband" & 354 & 0.397  \\
 & jnmom & ``just no mom", an annoying mother & 113 & 0.367  \\
 & justnos & annoying family members & 110 & 0.366 \\ 
 & jnso & ``just no significant other", an annoying romantic partner & 36 & 0.345   \\
 \hline
\multirow{ 4}{*}{r/gardening} & clematis & a type of flower & 150 & 0.395  \\
 & milkweed & a flowering plant & 156 & 0.389  \\
 & perennials & plants that live for multiple years & 139 & 0.383  \\ 
 & bindweed & a type of weed & 38 & 0.369  \\
 \hline
 \multirow{ 4}{*}{r/ps4} & siea & Sony Interactive Entertainment America & 60 & 0.373  \\
 & ps5 & PlayStation 5 & 892 & 0.371  \\ 
 & tlou & The Last of Us, a video game & 193 & 0.358 \\
 & hzd & Horizon Zero Dawn, a video game & 208 & 0.357 \\
\hline
\end{tabular}
}
\caption{\label{high_type_pmi} Examples of words with high type NPMI scores in three subreddits. We present values for this metric because as we will show in Section~\ref{sec:gloss}, it tends to perform better. The listed count is the number of unique users using that word in that subreddit. }
\end{table*}

\section{Data}\label{data}

Our data is a subset of all comments on Reddit made during May and June 2019 \cite{baumgartner2020pushshift}. Reddit is broken up into forum-based communities called \textit{subreddits}, which discuss different topics, such as parenting or gaming, or target users in different social groups, such as LGBTQ+ or women. We select the top 500 most popular subreddits based on number of comments and remove subreddits that have less than 85\% English comments, using the language identification method proposed by \citet{lui-baldwin-2012-langid}. This process yields 474 subreddits, from which we randomly sample 80,000 comments each. The number of comments per subreddit originally ranged from over 13 million to over 80 thousand, so this sampling ensures that more popular communities do not skew comparisons of word usage across subreddits. Each sampled subreddit had around 20k unique users on average, where a user is defined as a unique username associated with comments.\footnote{Some Reddit users may have multiple usernames due to the creation of ``throwaway" accounts \cite{10.1145/2675133.2675175}, but we define a single user by its account username.} We lowercase the text, remove urls, and replace usernames, numbers, and subreddit names each with their own special token type. The resulting dataset has over 1.4 billion tokens.

To understand how users in these communities define and catalog their own language, we also manually gather all available glossaries of the subreddits in our dataset. These glossaries are usually written as guides to newcomers to the community and can be found in or linked from community wiki pages. We exclude glossary links that are too general and not specific to that Reddit community, such as r/tennis's link to the Wikipedia page for tennis terms. We provide the names of these communities and the links we used in our Github repo.\footnote{\url{https://github.com/lucy3/ingroup_lang}} Our 57 subreddit glossaries have an average of 72.4 terms per glossary, with a wide range from a minimum of 4 terms to a maximum of 251. We removed 1044 multi-word expressions from analysis, because counting phrases would conflate the distinction we make between examining which individual words are used (type) and how they are used (meaning). We evaluate on 2814 single-token words from these glossaries that appear in comments within their respective subreddits based on exact string matching. Since many of these words appear in multiple subreddits' glossaries, we have 2226 unique glossary words overall. 

\section{Methods for Identifying Community-Specific Language}

\subsection{Type}

Much previous work on Internet language has focused on lexical choice, examining the word types unique to a community. The subreddit r/vegan, for example, uses \textit{carnis}, \textit{omnis}, and \textit{omnivores} to refer to people who eat meat. 

For our type-based analysis, we only examine words that are within the 20\% most frequent in a subreddit; even though much of a community's unique language is in its long tail, words with fewer than 10 occurrences may be noisy misspellings or too rare for us to confidently determine usage patterns. To keep our vocabularies compatible with our sense-based method described in \S\ref{s:meaning}, we calculate word frequencies using the basic (non-WordPiece) tokenizer in Hugging Face's transformers library\footnote{https://huggingface.co/transformers/} \cite{wolf-etal-2020-transformers}. Following \citet{eisenstein2014diffusion}, we define frequency for a word $t$ in a subreddit $s$, $f_s(t)$, as the number of users that used it at least once in the subreddit. We experiment with several different methods for finding distinctive and salient words in subreddits.

Our first metric is the ``specificity'' metric used in \citet{zhang2017community} to measure the distinctiveness of words in a community. For each word type $t$ in subreddit $s$, we calculate its PMI $\mathcal{T}$, which we will refer to as \textit{type PMI}: $$\mathcal{T}_s(t) = \log{P(t \mid s) \over P(t)}.$$ $P(t \mid s)$ is the probability of word $t$ in subreddit $s$, or $$P(t \mid s) = \frac{f_s(t)}{\sum_w{f_s(w)}},$$ while $P(t)$ is the probability of the word overall, or $$P(t) = \frac{\sum_r f_r(t)}{\sum_{w,r} f_r(w)}.$$ 

PMI can be normalized to have values between $[-1, 1]$, which also reduces its tendency to overemphasize low frequency events \cite{bouma2009normalized}. Therefore, we also calculate words' NPMI $\mathcal{T^\ast}$, or \textit{type NPMI}: $$\mathcal{T}^\ast_s(t) = \frac{\mathcal{T}_s(t)}{-\log{P(t, s)}}.$$ Here, $$P(t, s) = \frac{f_s(t)}{\sum_{w,r} f_r(w)}.$$

Table \ref{high_type_pmi} shows example words with high NPMI in three subreddits. The community r/justnomil, whose name means ``just no mother-in-law'', discusses negative family relationships, so many of its common and distinctive words refer to relatives. Words specific to other communities tend to be topical as well. The gaming community r/ps4 (PlayStation 4) uses acronyms to denote company and game entities and r/gardening has words for different types of plants. 

We also calculate term frequency–inverse document frequency (tf-idf) as a third alternative metric \cite{manning2008introduction}: $$\textrm{TFIDF}_s(t) = (1 + \log{f_s(t)})\log_{10}{\frac{N}{d(t)}},$$ where $N$ is the number of subreddits (474) and $d(t)$ is the number of subreddits word $t$ appears in. 

As another metric, we examine the use of \mbox{TextRank}, which is commonly used for extracting keywords from documents \cite{mihalcea-tarau-2004-textrank}. TextRank applies the PageRank algorithm \cite{brin1998anatomy} on a word co-occurrence graph, where the resulting scores based on words' positions in the graph correspond their importance in a document. For our use case, we construct a graph of unlemmatized tokens using the same parameter and model design choices as \citet{mihalcea-tarau-2004-textrank}. This means we run PageRank on an unweighted, undirected graph of adjectives and nouns that co-occur in the same comment, using a window size of 2, a convergence threshold of 0.0001, and a damping factor of 0.85. 

Finally, we also use Jensen-Shannon divergence (JSD), which has been used to identify divergent keywords in corpora such as books and social media \cite{Lin1991jsd, 10.1371/journal.pone.0195644, 10.1371/journal.pone.0137041, lu-etal-2020-diverging}. JSD is a symmetric version of Kullback–Leibler divergence, and it is preferred because it avoids assigning infinite values to words that only appear in one corpus. For each subreddit $s$, we compare its word probability distribution against that of a background corpus $R_s$ containing all other subreddits in our dataset. For each token $t$ in $s$, we calculate its divergence contribution as 
\begin{equation*}
\begin{split}
D_s(t) & = -m_s(t) \log_2{m_s(t)} \\ 
& + \frac{1}{2}(P(t \mid s) \log_2{P(t \mid s)} \\
& + P(t \mid R_s) \log_2{P(t \mid R_s)}),
\end{split}
\end{equation*}
where $$m_s(t) = \frac{P(t \mid s) + P(t \mid R_s)}{2}$$ \cite{lu-etal-2020-diverging, 10.1371/journal.pone.0137041}. Divergence scores are positive, and the computed score does not indicate in which corpus, $s$ or $R_s$, a word is more prominent. Therefore, we label $D_s(t)$ as negative if $t$'s contribution comes from $R_s$, or if $P(t \mid s) < P(t \mid R_s)$. 

\subsection{Meaning}\label{s:meaning}

Some words may have low scores with our type-based metrics, yet their use should still be considered community-specific. For example, the word \textit{ow} is common to many subreddits, but is used as an acronym for a video game name in r/overwatch, a clothing brand in r/sneakers, and how much a movie makes in its opening weekend in r/boxoffice. We use interpretable metrics for senses, analogous to type NPMI, that allow us to compare semantic variation across communities. 

Since words on social media are dynamic and niche, making them difficult to be comprehensively cataloged, we frame our task as word sense induction. We investigate two types of methods: one that clusters BERT \textbf{embeddings}, and \citet{amrami2019towards}'s current state-of-the-art model that clusters representatives containing word \textbf{substitutes} predicted by BERT (Figure~\ref{fig:fig_1}). 

The current state-of-the-art WSI model associates each example of a target word with 15 representatives, each of which is a vector composed of 20 sampled substitutes for the masked target word \cite{amrami2019towards}. This method then transforms these sparse vectors with tf-idf and clusters them using aggolomerative clustering, dynamically merging less probable senses with more dominant ones. In our use of this model, each example is assigned to its most probable sense based on how its representatives are distributed across sense clusters. One version of their model uses Hearst-style patterns such as \textit{target (or even [MASK])}, instead of simply masking out the target word. We do not use dynamic patterns in our study, because these patterns assume that target words are nouns, verbs, or adjectives, and our Reddit experiments do not filter out any words based on part of speech. 

As we will show, \citet{amrami2019towards}'s model is resource intensive on large datasets, and so we also test a more lightweight method that has seen prior application on similar tasks. Pre-trained BERT-base\footnote{We also experimented with a BERT model after domain-adaptive pretraining on our entire Reddit dataset \cite{han-eisenstein-2019-unsupervised,gururangan-etal-2020-dont}, and reached similar results in our Reddit language analyses.} has demonstrated good performance on word sense disambiguation and identification using embedding distance-based techniques \cite{wiedemann-etal-2019-does, hu2019diachronic,NIPS2019_9065,hadiwinoto-etal-2019-improved}. The positions of dimensionality-reduced BERT representations for \textit{python} in Figure~\ref{fig:fig_1} suggest that they are grouped based on their community-specific meaning. Our embedding-based method discretizes these hidden layer landscapes across hundreds of communities and thousands of words. This method is $k$-means \cite{kmeans1982, 10.5555/1283383.1283494, scikit-learn}, which has also been employed by concurrent work to track word usage change over time \cite{giulianelli-etal-2020-analysing}. We cluster on the concatenation of the final four layers of BERT.\footnote{We also tried other ways of forming embeddings, such as summing all layers \cite{giulianelli-etal-2020-analysing}, only taking the last layer \cite{hu2019diachronic}, and averaging all layers, but concatenating the last four performed best.} There have been many proposed methods for choosing $k$ in $k$-means clustering, and we experimented with several of these, including the gap statistic \cite{tibshirani2001estimating} and a variant of $k$-means using the Bayesian information criterion (BIC) called $x$-means \cite{pelleg2000x}. The following criterion for cluster cardinality worked best on development set data \cite{manning2008introduction}: $$k = \textrm{argmin}_k \textrm{RSS}(k) + \gamma k,$$ where $\textrm{RSS}(k)$ is the minimum residual sum of squares for number of clusters $k$ and $\gamma$ is a weighting factor. 

We also tried applying spectral clustering on BERT embeddings as a possible alternative to $k$-means \cite{spectral2000, VonLuxburg2007, scikit-learn}. Spectral clustering turns the task of clustering embeddings into a connectivity problem, where similar points have edges between them, and the resulting graph is partitioned so that points within the same group are similar to each other, while those across different groups are dissimilar. To do this, $k$-means is not applied directly on BERT embeddings, but instead on a projection of the similarity graph's normalized Laplacian. We use the nearest neighbors approach for creating the similarity graph, as recommended by \citet{VonLuxburg2007}, since this construction is less sensitive to parameter choices than other graphs. To determine the number of clusters $k$, we used the eigengap heuristic: $$k = \text{argmax}_k{\lambda_{k+1} - \lambda_k},$$ where $\lambda_k$ for $k = 1, ..., 10$ are the smallest eigenvalues of the similarity graph's normalized Laplacian. 

\section{Word Sense Induction}

We develop and evaluate word sense induction models using \mbox{SemEval} WSI tasks in a manner that is designed to parallel their later use on larger Reddit data.

\subsection{Evaluation on SemEval Tasks}

In \mbox{SemEval} 2010 Task 14 \cite{jurgens-klapaftis-2013-semeval} and \mbox{SemEval} 2013 Task 13 \cite{manandhar-klapaftis-2009-semeval}, models are evaluated based on how well predicted sense clusters for different occurrences of a target word align with gold sense clusters.

\citet{amrami2019towards}'s performance scores reported in their paper were obtained from running their model directly on test set data for the two \mbox{SemEval} tasks, which had typically fewer than 150 examples per word. However, these tasks were released as multi-phase tasks and provide both training and test sets \cite{jurgens-klapaftis-2013-semeval,manandhar-klapaftis-2009-semeval}, and our study requires methods that can scale to larger datasets. Some words in our Reddit data appear very frequently, making it too memory-intensive to cluster all of their embeddings or representatives at once (for example, the word \textit{pass} appears over 96k times). It is more feasible to learn senses from a fixed number of examples, and then match remaining examples to these senses. We evaluate how well induced senses generalize to new examples using separate train and test sets. 

We tune parameters for models using \mbox{SemEval} 2010 Task 14. In this task, the test set contains 100 target noun and verb lemmas, where each occurrence of a lemma is labeled with a single sense \cite{manandhar-klapaftis-2009-semeval}. We use WSI models to first induce senses for 500 randomly sampled training examples, and then match test examples to these senses. There are a few lemmas in \mbox{SemEval} 2010 that occur fewer than 500 times in the training set, in which case we use all instances. We also evaluate the top-performing versions of each model on \mbox{SemEval} 2013 Task 13, after clustering 500 instances of each noun, verb, or adjective lemma in their training corpus, \mbox{ukWaC} \cite{jurgens-klapaftis-2013-semeval,baroni2009wacky}. In \mbox{SemEval} 2013 Task 13, each occurrence of a word is labeled with multiple senses, but we evaluate and report past scores using their single-sense evaluation key, where each word is mapped to one sense.

For the substitution-based method, we match test examples to clusters by pairing representatives with the sense label of their nearest neighbor in the training set. We found that \citet{amrami2019towards}'s default model is sensitive to the number of examples clustered. The majority of target words in the test data for the two \mbox{SemEval} tasks on which this model was developed have fewer than 150 examples. When this same model is applied on a larger set of 500 examples, the vast majority of examples often end up in a single cluster, leading to low or zero-value V-Measure scores for many words. To mitigate this problem, we experimented with different values for the upper-bound on number of clusters $c$, ranging from 10 to 35 in increments of 5. This upper-bound determines the distance threshold for flattening dendrograms, where allowing more clusters lowers these thresholds and breaks up large clusters. We found $c=25$ produces the best \mbox{SemEval} 2010 results for our training set size, and use it for our Reddit experiments as well. 

For the $k$-means embedding-based method, we match test examples to the nearest centroid representing an induced sense using cosine distance. During training, we initialize centroids using $k$-means++ \cite{10.5555/1283383.1283494}. We experimented with different values of the weighting factor $\gamma$ ranging from 1000 to 20000 on \mbox{SemEval} 2010, and choose $\gamma=10000$ for our experiments on Reddit data. Preliminary experiments suggest that this method is less sensitive to the number of training examples, where directly clustering \mbox{SemEval} 2010's smaller test set led to similar results with the same parameters. 

For the spectral embedding-based method, we match a test example to a cluster by assigning it the label of its nearest training example. To construct the $K$-nearest neighbor similarity graph during training, we experimented with different $K$ around $\log(n)$, where for $n=500$, $K \sim 6$ \cite{VonLuxburg2007, BRITO199733}. For $K = 6, ..., 10$, we found that $K=7$ worked best, though performance scores on \mbox{SemEval} 2010 for all other $K$ were still within one standard deviation of $K=7$'s average across multiple runs. 

The bolded rows of Table \ref{semeval2010} and Table \ref{semeval2013} show performance scores of these models using our evaluation setup, compared against scores reported in previous work.\footnote{The single-sense scores for \citet{amrami2019towards} are not reported in their paper. To generate these scores, we ran the default model in their code base directly on the test set using \mbox{SemEval} 2013's single-sense evaluation key, reporting average performance over ten runs.} These results show that for embedding-based WSI, $k$-means works better than spectral clustering. In addition, clustering BERT embeddings performs better than most methods, but not as well as clustering substitution-based representatives.

\begin{table}[t]
\centering
\tiny
\resizebox{\columnwidth}{!}{%
\begin{tabular}{>{\raggedright}p{2.4cm} | ccc}
\hline \textbf{Model} & \textbf{F Score} & \textbf{V Measure} & \textbf{Average} \\ \hline
\textbf{BERT embeddings}, $k$-means, $\gamma=10000$ & 0.594 (0.004) & 0.306 (0.004) & 0.426 (0.003)\\\hline
\textbf{BERT embeddings}, spectral, $K=7$ & 0.581 (0.025) & 0.283 (0.017) & 0.405 (0.020)\\\hline
\textbf{BERT substitutes}, \citet{amrami2019towards}, $c=25$ & 0.683 (0.003) & 0.339 (0.012)  & 0.481 (0.009)\\\hline
\citet{amrami2019towards}, default parameters & 0.709 & 0.378 & 0.517 \\\hline
\citet{amplayo2019autosense} & 0.617 & 0.098 & 0.246 \\\hline
\citet{song2016sense} & 0.551 & 0.098 & 0.232 \\\hline
\citet{chang-etal-2014-inducing} & 0.231 & 0.214 & 0.222 \\\hline
MFS & 0.635 & 0.000 & 0.000 \\ \hline
\end{tabular}
}
\caption{\label{semeval2010} SemEval 2010 Task 14 unsupervised evaluation results with two measures, F Score and V Measure, and their geometric mean. MFS is most frequent sense baseline, where all instances are assigned to the most frequent sense. Standard deviation over five runs are in parentheses. Bolded models use our train and test evaluation setup.}
\end{table}

\begin{table}[t]
\centering
\tiny
\resizebox{\columnwidth}{!}{%
\begin{tabular}{>{\raggedright}p{2.4cm} | ccc}
\hline \textbf{Model} & \textbf{NMI} & \textbf{B-Cubed} & \textbf{Average} \\ \hline
\textbf{BERT embeddings}, $k$-means, $\gamma=10000$ & 0.157 (0.006) &  0.575 (0.005) & 0.300 (0.007) \\\hline
\textbf{BERT embeddings}, spectral, $K=7$ & 0.135 (0.010) & 0.588 (0.007) & 0.282 (0.010) \\\hline
\textbf{BERT substitutes}, \citet{amrami2019towards}, $c=25$ & 0.192 (0.011) & 0.638 (0.003)  & 0.350 (0.010) \\\hline
\citet{amrami2019towards}, default parameters & 0.183 & 0.626 & 0.339\\\hline
\citet{baskaya-etal-2013-ai} &  0.045 & 0.351 & 0.126 \\\hline
\citet{lau2013unimelb} &  0.039 & 0.441 & 0.131 \\\hline
\end{tabular}
}
\caption{\label{semeval2013} SemEval 2013 Task 13 single-sense evaluation results with two measures, NMI and B-Cubed, and their geometric mean. Standard deviation over five runs are in parentheses. Bolded models use our train and test evaluation setup.}
\end{table}

\begin{table}[t]
\centering
\tiny
\resizebox{\columnwidth}{!}{%
\begin{tabular}{>{\raggedright}p{2.1cm} | p{1.5cm}p{1.5cm}}
\hline \textbf{Model} & \textbf{Clustering per word} & \textbf{Matching per subreddit} \\ \hline
\textbf{BERT embeddings}, $\gamma=10000$ & 47.60 sec & 28.85 min \\\hline
\textbf{\citet{amrami2019towards}'s BERT substitutes}, $c=25$ & 80.99 sec & 23.04 hr \\
\hline
\end{tabular}
}
\caption{\label{reddit_timing} The models' median time clustering 500 examples of each word, and their median time matching all words in a subreddit to senses.}
\end{table}

\begin{table*}[t]
\centering
\tiny
\resizebox{\textwidth}{!}{%
\begin{tabular}{l | ccccp{2.5cm}p{2.5cm}}
\hline \textbf{subreddit} & \textbf{word} & \textbf{$\mathcal{M}^\dagger$} & \textbf{\tiny$\mathcal{M}^\ast$} & \textbf{$\mathcal{T^\ast}$} & \textbf{subreddit's sense example} & \textbf{other sense  example} \\ \hline
r/elitedangerous & python & 0.383 & 0.347 & 0.286 & ``Get a \textbf{Python}, stuff it with passenger cabins..." & ``I self taught some \textbf{Python} over the summer..." \\ \hline
r/fashionreps & haul & 0.374 & 0.408 & 0.358 & ``Plan your first \textbf{haul}, don't just buy random nonsense..." & ``...discipline is the long \textbf{haul} of getting it done..." \\ \hline
r/libertarian & nap & 0.370 & 0.351 & 0.185 & ``The \textbf{nap} is just a social contract." & ``Move bedtime earlier to compensate for no \textbf{nap}..." \\ \hline
r/90dayfiance & nickel & 0.436 & 0.302 & 0.312 & ``\textbf{Nickel} really believes that Azan loves her." & ``...raise burrito prices by a \textbf{nickel} per month..." \\ \hline
r/watches & dial & 0.461 & 0.463 & 0.408 & ``...the \textbf{dial} has a really nice texturing..." & ``...you didn't have to \textbf{dial} the area code..." \\
\hline
\end{tabular}
}
\caption{\label{high_sense_pmi} Examples of words where both the embedding-based and substitution-based WSI models result in a high sense NPMI score in the listed subreddit. Each row includes example contexts from comments illustrating the subreddit-specific sense and a different sense pulled from a different subreddit.}
\end{table*}

\subsection{Adaptation to Reddit}

We apply the $k$-means embedding-based method and \citet{amrami2019towards}'s substitution-based method to Reddit, with the parameters that performed best on \mbox{SemEval} 2010 Task 14. We induce senses for a vocabulary of non-lemmatized 13,240 tokens, including punctuation, that occur often enough for us to gain a strong signal of semantic deviation from the norm. These are non-emoji tokens that are very common in a community (in the top 10\% most frequent tokens of a subreddit), frequent enough to be clustered (appear at least 500 times overall), and also used broadly (appear in at least 350 subreddits). When clustering BERT embeddings, to gain the representation for a token split into wordpieces, we average their vectors. With each WSI method, we induce senses using 500 randomly sampled comments containing the target token.\footnote{To avoid sampling repeated comments written by bots, we disregarded comments where the context window around a target word (five tokens to the left and five tokens to the right) repeat 10 or more times.} Then, we match all occurrences of words in our selected vocabulary to their closest sense, as described earlier. 

Though the embedding-based method has lower performance than the substitution-based one on \mbox{SemEval} WSI tasks, the former is an order of magnitude faster and more efficient to scale (Table~\ref{reddit_timing}).\footnote{We used a Tesla K80 GPU for the majority of these experiments, but we used a TITAN Xp GPU for three of the 474 subreddits for the substitution-based method.} 
During the training phase of clustering, both models learn sense clusters for each word by making a single pass over that word's set of examples; we then match every vocab word in a subreddit to its appropriate cluster. 
While the substitution-based method is 1.7 times slower than the embedding-based method during the training phase, it becomes 47.9 times slower during the matching phase.
The particularly large difference in runtime  is due to the substitution-based method's need to run BERT \emph{multiple} times for each sentence (in order to individually mask each vocab word in the sentence),
while the embedding-based method passes over each sentence once. We also noticed that the substitution-based method sometimes created very small clusters, which often led to very rare senses (e.g. occurring fewer than 5 times overall). 

After assigning words to senses using a WSI model, we calculate the NPMI of a sense $n$ in subreddit $s$, counting each sense once per user: $$\mathcal{S}_s(n) = \log{P(n \mid s) \over P(n)} \bigg/ -\log{P(n, s)},$$ where $P(n \mid s)$ is the probability of sense $n$ in subreddit $s$, $P(n, s)$ is the joint probability of $n$ and $s$, and $P(n)$ is the probability of sense $n$ overall. 

A word may map to more than one sense, so to determine if a word $t$ has a community-specific sense in subreddit $s$, we use the NPMI of the word's most common sense in $s$. We refer to this value as the \textit{sense NPMI}, or $\mathcal{M}_s(t)$. We calculate these scores using both the embedding-based method, denoted as $\mathcal{M}^\ast_s(t)$, and the substitution-based method, denoted as $\mathcal{M}^\dagger_s(t)$.

These two sense NPMI metrics tend to score words very similarly across subreddits, with an overall Pearson's correlation of 0.921 ($p < 0.001$). Words that have high NPMI with one model also tend to have high NPMI with the other (Table~\ref{high_sense_pmi}). There are some disagreements, such as the scores for \textit{flu} in r/keto, which does not refer to influenza but instead refers to symptoms associated with starting a ketogenic diet ($\mathcal{M}^\ast = 0.388$, $\mathcal{M}^\dagger = 0.248$). Still, both metrics place r/keto's \textit{flu} in the 98th percentile of scored words. Thus, for large datasets, it would be worthwhile to use the embedding-based method instead of the state-of-the-art substitution-based method to save substantial time and computing resources and yield similar results. 

Some of the words with high sense NPMI in Table \ref{high_sense_pmi}, such as \textit{haul} (a set of purchased products), \textit{dial} (a watch face) have well documented meanings in WordNet or the \emph{Oxford English Dictionary} that are especially relevant to the topic of the community. Others are less standard, including \textit{python} to refer to a ship in a game, \textit{nap} as an acronym for ``non‐aggression principle", and \textit{Nickel} as a fan-created nickname for a character named Nicole in a reality TV show. Some terms have low $\mathcal{M}$ across most subreddits, such as the period punctuation mark (average $\mathcal{M}^\ast = -0.008$, $\mathcal{M}^\dagger = -0.009$). 

\section{Glossary Analysis}\label{sec:gloss}

\begin{table*}
\centering
\tiny
\resizebox{\textwidth}{!}{%
\begin{tabular}{l|>{\raggedright}p{2.3cm} | p{1.1cm} | p{0.8cm}p{1.2cm} | p{1.4cm}p{1.8cm}}
\hline & \textbf{metric} & \textbf{mean \newline reciprocal rank} & \textbf{median, \newline glossary words} & \textbf{median, \newline non-glossary words} & \textbf{98th percentile, all words} & \textbf{\% of scored \newline glossary words in 98th percentile}\\ \hline
\multirow{5}{*}{\textbf{type}} & PMI ($\mathcal{T}$) & 0.0938 & 2.7539 & 0.2088 & 5.0063 & 18.13 \\
& NPMI ($\mathcal{T^\ast}$) & 0.4823 & 0.1793 & 0.0131 & 0.3035 & 22.30 \\
& TFIDF & 0.2060 & 0.5682 & 0.0237 & 3.0837 & 16.76\\ 
& TextRank & 0.0616 & 6.95e-5 & 7.90e-5 & 0.0002 & 24.91 \\
& JSD & 0.2644 & 2.02e-5 & 2.44e-7 & 5.60e-05 & 29.07 \\\hline
\multirow{2}{*}{\textbf{sense}} & BERT substitutes ($\mathcal{M}^\dagger$) & 0.2635 & 0.1165 & 0.0143 & 0.1745 &  28.75\\
& BERT embeddings ($\mathcal{M}^\ast$) & 0.3067 & 0.1304 & 0.0208 & 0.1799 & 30.73 \\\hline
\end{tabular}
}
\caption{\label{gloss_scores} This table compares how each metric for quantifying community-specific language handles words in user-created subreddit glossaries. The 98th percentile cutoff for all words are calculated for each metric using all scores across all subreddits. The \% of glossary words is based on the fraction of glossary words with calculated scores for each metric.}
\end{table*}

To provide additional validation for our metrics, we examine how they score words listed in user-created subreddit glossaries (as described in \S \ref{data}). New members may spend 8 to 9 months acquiring a community's linguistic norms \cite{nguyen2011language}, and some Reddit communities have such distinctive language that their posts can be difficult to understand to outsiders. This makes the manual annotation of linguistic norms across hundreds of communities difficult, and so for the purposes of our study, we use user-created glossaries to provide context for what our metrics find. Still, glossaries only contain words deemed by a few users to be important for their community, and the lack of labeled negative examples inhibits their use in a supervised machine learning task. Therefore, we focus on whether glossary words, on average, tend to have high scores using our methods. 

Table \ref{gloss_scores} shows that glossary words have higher median scores than non-glossary words for all listed metrics (U-tests, $p < 0.001$). In addition, a substantial percentage of glossary words are in the 98th percentile of scored words for each metric.  

To see how highly our metrics tend to score glossary terms, we calculate their mean reciprocal rank (MRR), an evaluation metric often used to evaluate query responses \cite{voorhees1999trec}: $$\textrm{mean reciprocal rank} = \frac{1}{G}\sum_{i = 1}^{G} \frac{1}{\textrm{rank}_i},$$ where $\textrm{rank}_i$ is the rank position of the highest scored glossary term for a subreddit and $G$ is the number of subreddits with glossaries. Mean reciprocal rank ranges from 0 to 1, where 1 would mean a glossary term is the highest scored word for all subreddits. 

We have five different possible metrics for scoring community-specific word types: type PMI, type NPMI, tf-idf, TextRank, and JSD. Of these, TextRank has the lowest MRR, but still scores a competitive percentage of glossary words in the 98th percentile. This is because the TextRank algorithm only determines how important a word is within each subreddit, without any comparison to other subreddits to determine how a word's frequency in a subreddit differs from the norm. Type NPMI has the highest MRR, followed by JSD. Though JSD has more glossary words in the 98th percentile than type NPMI, we notice that many high-scoring JSD terms include words that have a very different probability in a subreddit compared to the rest of Reddit, but are not actually distinctive to that subreddit. For example, in r/justnomil, words such as \textit{husband}, \textit{she}, and \textit{her} are within the top 10 ranked words by JSD score. This contrasts the words in Table~\ref{high_type_pmi} with high NPMI scores that are more unique to r/justnomil's vocabulary. Therefore, for the remainder of this paper, we focus on NPMI as our type-based metric for measuring lexical variation. 

\begin{figure}[t]
	\centering{\includegraphics[width=0.9\linewidth]{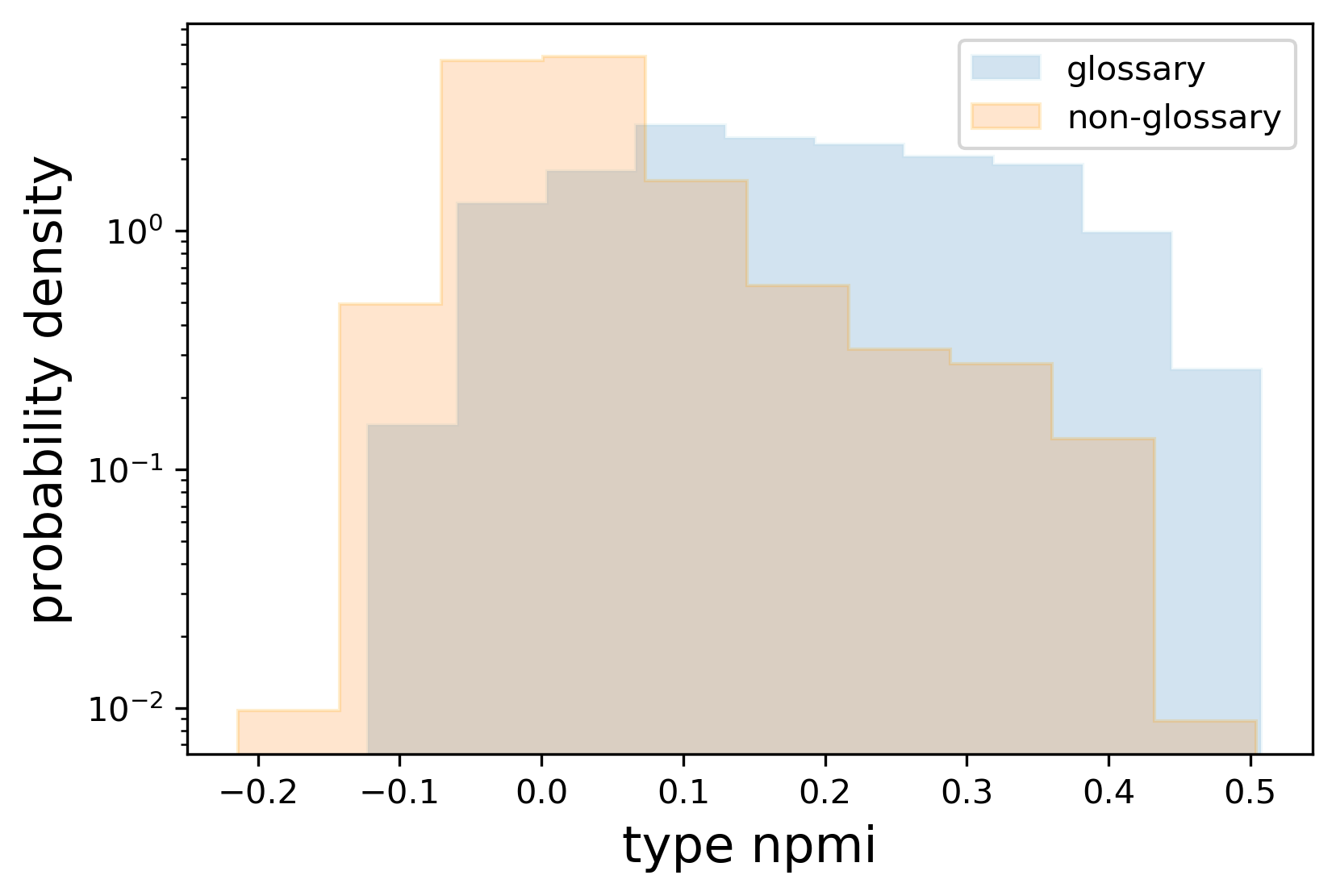}}
	\centering{\includegraphics[width=0.9\linewidth]{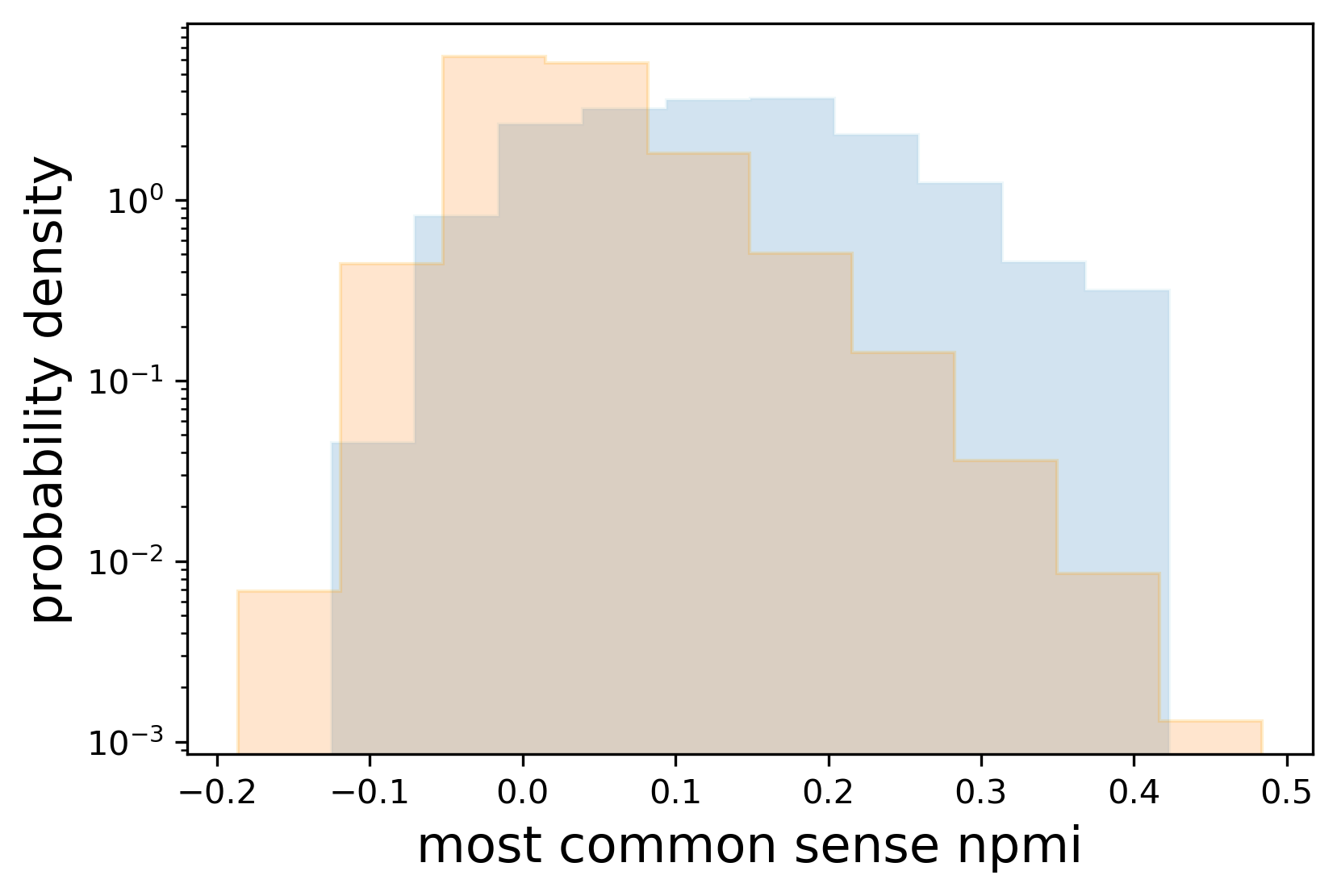}}
	\caption{Normalized distributions of type NPMI ($\mathcal{T}^\ast$) and sense NPMI ($\mathcal{M}^\ast$) for words in subreddits with user-created glossaries. The top graph involves 2,184 glossary words and 431,773 non-glossary words, and the bottom graph involves 807 glossary words and 194,700 non-glossary words. Glossary words tend to have higher scores than non-glossary words.} 
	\label{fig:gloss_pmi}
\end{figure}

Figure~\ref{fig:gloss_pmi} shows the normalized distributions of type NPMI and sense NPMI. Though glossary words tend to have higher NPMI scores than non-glossary words, there is still overlap between the two distributions, where some glossary words have low scores and some non-glossary words have high ones. Sometimes this is because many glossary words with low type NPMI instead have high sense NPMI. For example, the glossary word \textit{envy} in r/competitiveoverwatch refers to an esports team and has low type NPMI ($\mathcal{T^\ast}= 0.1876$) but sense NPMI in the 98th percentile ($\mathcal{M}^\ast = 0.2640$, $\mathcal{M}^\dagger = 0.2136$). Only 21 glossary terms, such as \textit{aha}, a popular type of skin exfoliant in r/skincareaddiction, are both in the 98th percentile of $\mathcal{T^\ast}$ and the 98th percentiles of $\mathcal{M}^\ast$ and $\mathcal{M}^\dagger$ scores. Thus, examining variation in the meaning of broadly used words provides a complementary metric to counting distinctive word types, and overall provides a more comprehensive understanding of community-specific language.

Other cases of overlap are due to model error. Manual inspection reveals that some glossary words that actually have unique senses have low $\mathcal{M}$ scores. Sometimes a WSI method splits a glossary term in a community into too many senses or fails to disambiguate different meanings. For example, the glossary word \textit{spawn} in r/childfree refers to children, but the embedding-based method assigns it to the same sense used in gaming communities, where it instead refers to the creation of characters or items. As another example of a failure case, the substitution-based method splits the majority of occurrences of \textit{rep}, an exercise movement, in r/bodybuilding into two large but separate senses. Though new methods using BERT have led to performance boosts, WSI is still a challenging task. 

The use of glossaries in our study has several limitations. Some non-glossary terms have high scores because glossaries are not comprehensive. For example, \textit{dips} ($\mathcal{M}^\ast = 0.2920$, $\mathcal{M}^\dagger = 0.2541$) is not listed in r/fitness's glossary, but it regularly refers to a type of exercise. This suggests the potential of our methods for uncovering possible additions to these glossaries. The vast majority of glossaries contain community-specific words, but a few also include common Internet terms that have low values across all metrics, such as \textit{lol}, \textit{imo}, and \textit{fyi}. In addition, only 71.12\% of all single-token glossary words occurred often enough to have scores calculated for them. Some words are relevant to the topic of the community (e.g. \textit{christadelphianism} in r/christianity), but are actually rarely used in discussions. We do not compute scores for rarely-occurring words, so they are excluded from our results. Despite these limitations, however, user-created glossaries are valuable resources for outsiders to understand the terminology used in niche online communities, and offer one of the only sources of in-domain validation for these methods.

\section{Communities and Variation}\label{sociolects}

In this section, we investigate how language variation relates to characteristics of users and communities in our dataset. For these analyses, we use the metrics that aligned the most with user-created glossaries (Table~\ref{gloss_scores}): $\mathcal{T^\ast}$ for lexical variation and $\mathcal{M}^\ast$ for semantic variation. We define $\mathcal{F}$, or the distinctiveness of a community's language variety, as the fraction of unique words in the community's top 20\% most frequent words that have $\mathcal{T^\ast}$ or $\mathcal{M}^\ast$ in the 98th percentile of all scores for each metric. That is, a word in a community is counted as a ``community-specific word'' if its $\mathcal{T^\ast} > 0.3035$ or if its $\mathcal{M}^\ast > 0.1799$. Though in the following subsections we report numerical results using these cutoffs, the U-tests for community-level attributes and $\mathcal{F}$ are statistically significant ($p < 0.0001$) for cutoffs as low as the 50th percentile. 

\subsection{User Behavior}

Online communities differ from those in the offline world due to increased anonymity of the speakers and a lack of face-to-face interactions. However, the formation and survival of online communities still tie back to social factors. One central goal of our work is to see what behavioral characteristics a community with unique language tends to have. We examine four user-based attributes of subreddits: community size, user activity, user loyalty, and network density. We calculate values corresponding to these attributes using the entire, unsampled dataset of users and comments. For each of these user-based attributes, we propose and test hypotheses on how they relate to how much a community's language deviates from the norm. Some of these hypotheses are pulled from established sociolinguistic theories previously developed using offline communities and interactions, and we test their conclusions in our large-scale, digital domain. We construct U-tests for each attribute after $z$-scoring them across subreddits, comparing subreddits separated into two equal-sized groups of high and low $\mathcal{F}$. 

\begin{figure}[t]
	\centering{\includegraphics[height=0.4\columnwidth]{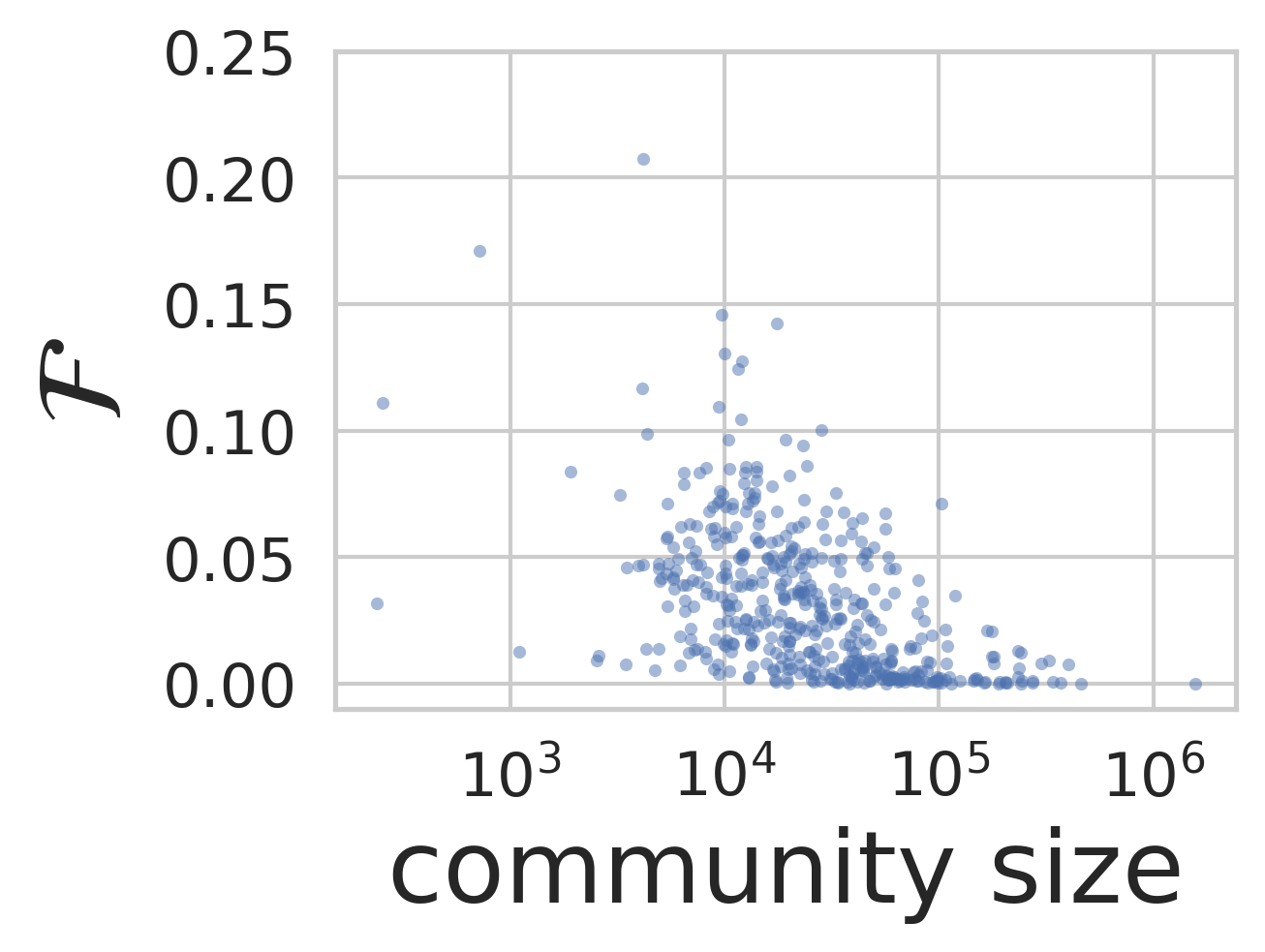}}
	\centering{\includegraphics[height=0.391\columnwidth]{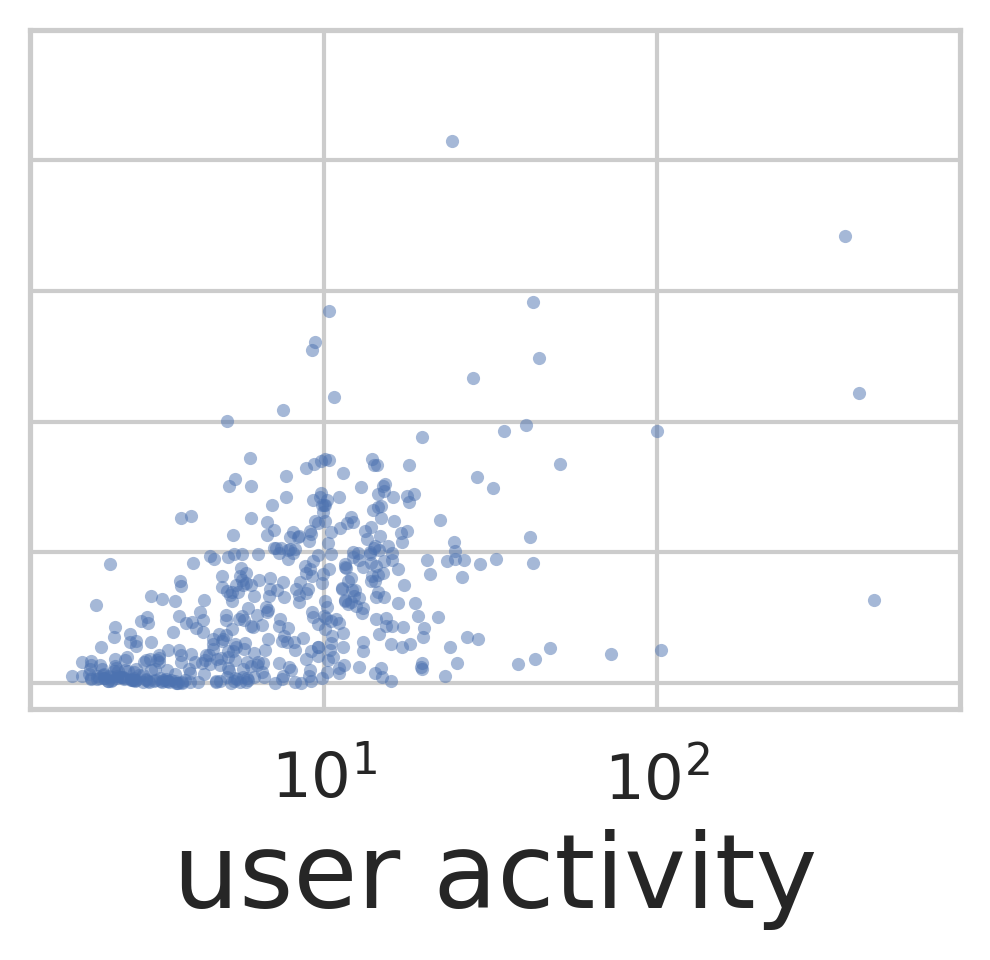}}
	\centering{\includegraphics[height=0.4\columnwidth]{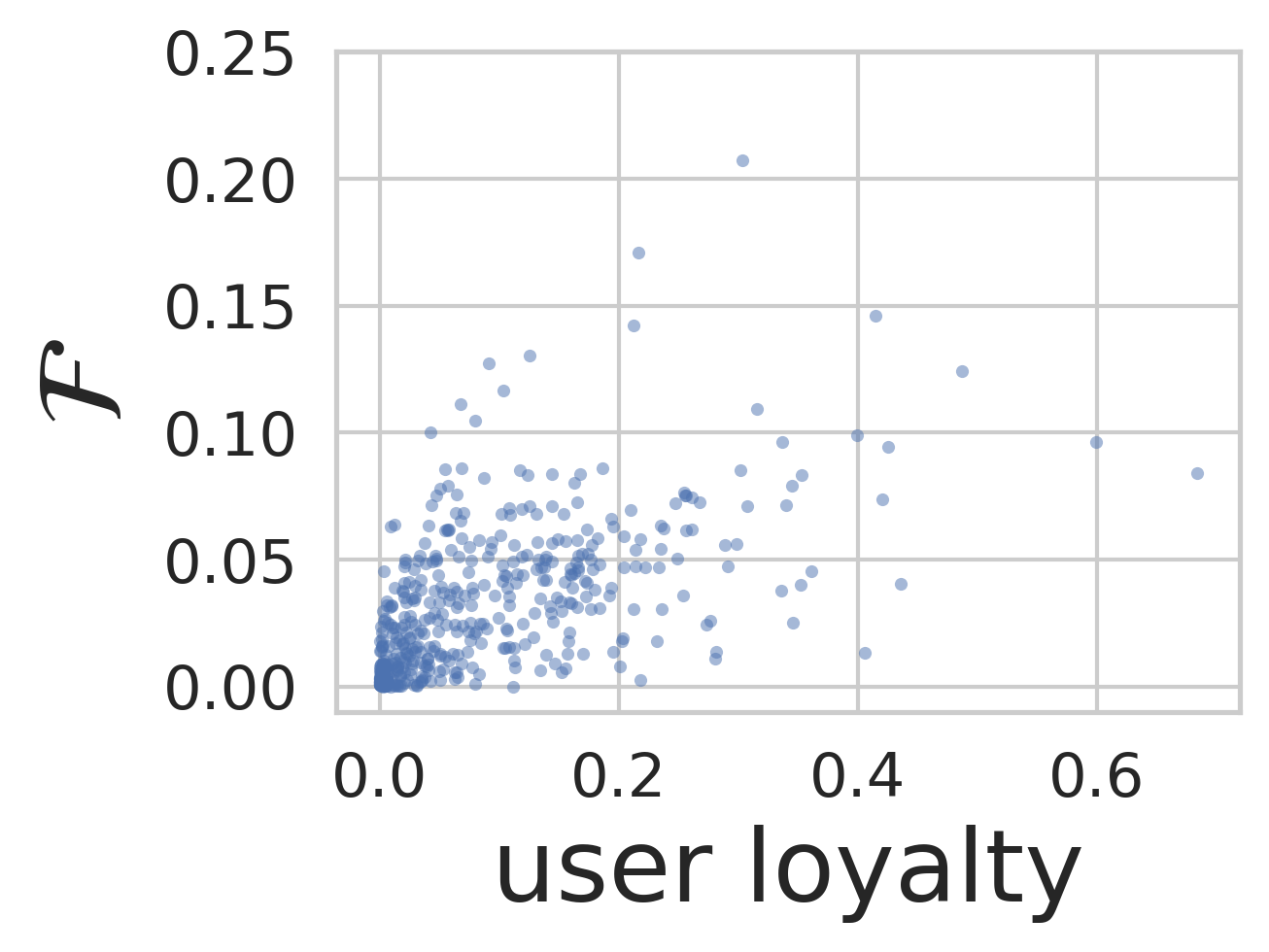}}
	\centering{\includegraphics[height=0.391\columnwidth]{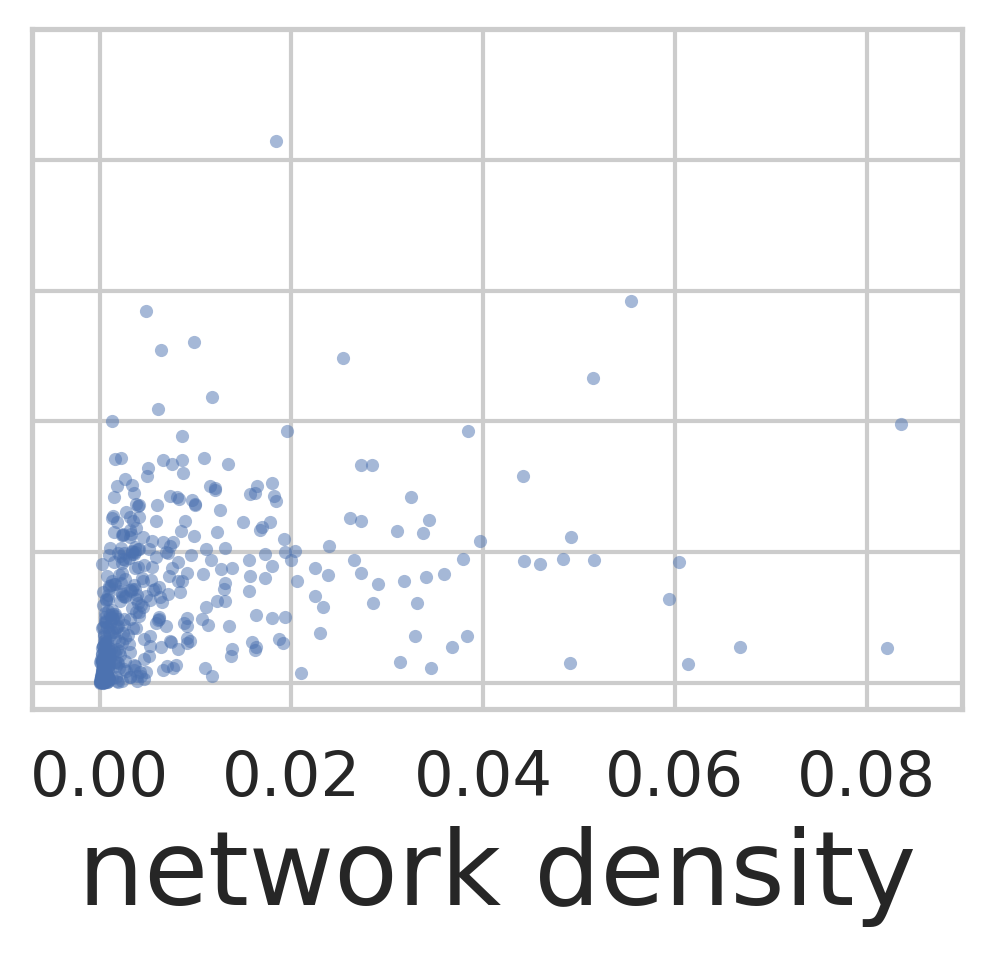}}
	\caption{Community size, user activity, user loyalty, network density all relate to the distinctiveness of a community's language, which is the fraction of words with type NPMI or sense NPMI scores in the 98th percentile. Each point on each plot represents one Reddit community. For clarity, axis limits are slightly cropped to omit extreme outliers.} 
	\label{fig:users}
\end{figure}

\citet{del-tredici-fernandez-2018-road}, when choosing communities for their study, claim that ``small-to-medium sized'' communities would be more likely to have lexical innovations. We define \textbf{community size} to be the number of unique users in a subreddit, and find that large communities tend to have less community-specific language ($p < 0.001$, Figure~\ref{fig:users}). Communities need to reach a ``critical mass'' to sustain meaningful interactions, but very large communities such as r/askreddit and r/news may suffer from communication overload, leading to simpler and shorter replies by users and fewer opportunities for group identity to form \cite{jones2004information}. We also collected subscriber counts from the last post of each subreddit made in our dataset's timeframe, and found that communities with more subscribers have lower $\mathcal{F}$ ($p < 0.001$), and communities with a higher ratio of subscribers to commenters also have lower $\mathcal{F}$ ($p < 0.001$). Multiple subreddits were outliers with extremely large subscriber counts, perhaps due to past users being auto-subscribed to default communities or historical popularity spikes. Future work could look into more refined methods of estimating the number of users who browse but do not comment in communities \cite{sun2014understanding}.

Active communities of practice require regular interaction among their members \cite{holmes1999community, wenger1998communities}. Our metric for measuring \textbf{user activity} is the average number of comments per user in that subreddit, and we find that communities with more community-specific language have more active users ($p < 0.001$, Figure \ref{fig:users}). However, within each community, we did not find significant or meaningful correlations between a user's number of comments in that community and the probability of them using a community-specific word.

Speakers with more local engagement tend to use more vernacular language, as it expresses local identity \cite{eckert2012three, bucholtz2005identity}. Our proxy for measuring this kind of engagement is the fraction of loyal users in a community, where loyal users are those who have at least 50\% of their comments in that particular subreddit. We use the definition of \textbf{user loyalty} introduced by \citet{hamilton2017loyalty}, filtering out users with fewer than 10 comments and counting only top-level comments. Communities with more community-specific language have more loyal users, which extends \citet{hamilton2017loyalty}'s conclusion that loyal users value collective identity ($p < 0.001$, Figure \ref{fig:users}). We also found that in 93\% of all communities, loyal users had a higher probability of using a word with $\mathcal{M}^\ast$ in the 98th percentile than a nonloyal user (U-test, $p < 0.001$), and in 90\% of all communities, loyal users had a higher probability of using a word with $\mathcal{T}^\ast$ in the 98th percentile (U-test, $p < 0.001$). Thus, users who use Reddit mostly to interact in a single community demonstrate deeper acculturation into the language of that community. 

A speech community is driven by the density of its communication, and dense networks enforce shared norms \cite{guy_2011, milroy1992social, sharma2020language}. Previous studies of face-to-face social networks may define edges using friend or familial ties, but Reddit interactions can occur between strangers. For \textbf{network density}, we calculate the density of the undirected direct-reply network of a subreddit based on comment threads: an edge exists between two users if one replies to the other. Following \citet{hamilton2017loyalty}, we only consider the top 20\% of users when constructing this network. More dense communities exhibit more community-specific language ($p < 0.001$, Figure \ref{fig:users}). Previous work using ethnography and friendship naming data has shown that a speaker's position in a social network is sometimes reflected in the language they use, where individuals on the periphery adopt less of the vernacular of a social group compared to those in the core \cite{labov1973linguistic, milroy1987language,sharma2020language}. To see whether users' position in Reddit direct-reply networks show a similar phenomena, we use \citet{cohen2014computing}'s method to approximate users' closeness centrality ($\epsilon = 10^{-7}$, $k=5000$). Within each community, we did not find a meaningful correlation between closeness centrality and the probability of a user using a community-specific word. This finding suggests that conversation networks on Reddit may not convey a user's degree of belonging to a community in the same manner as relationship networks in the physical world. 

\begin{figure}[t]
	\centering{\includegraphics[width=0.8\linewidth]{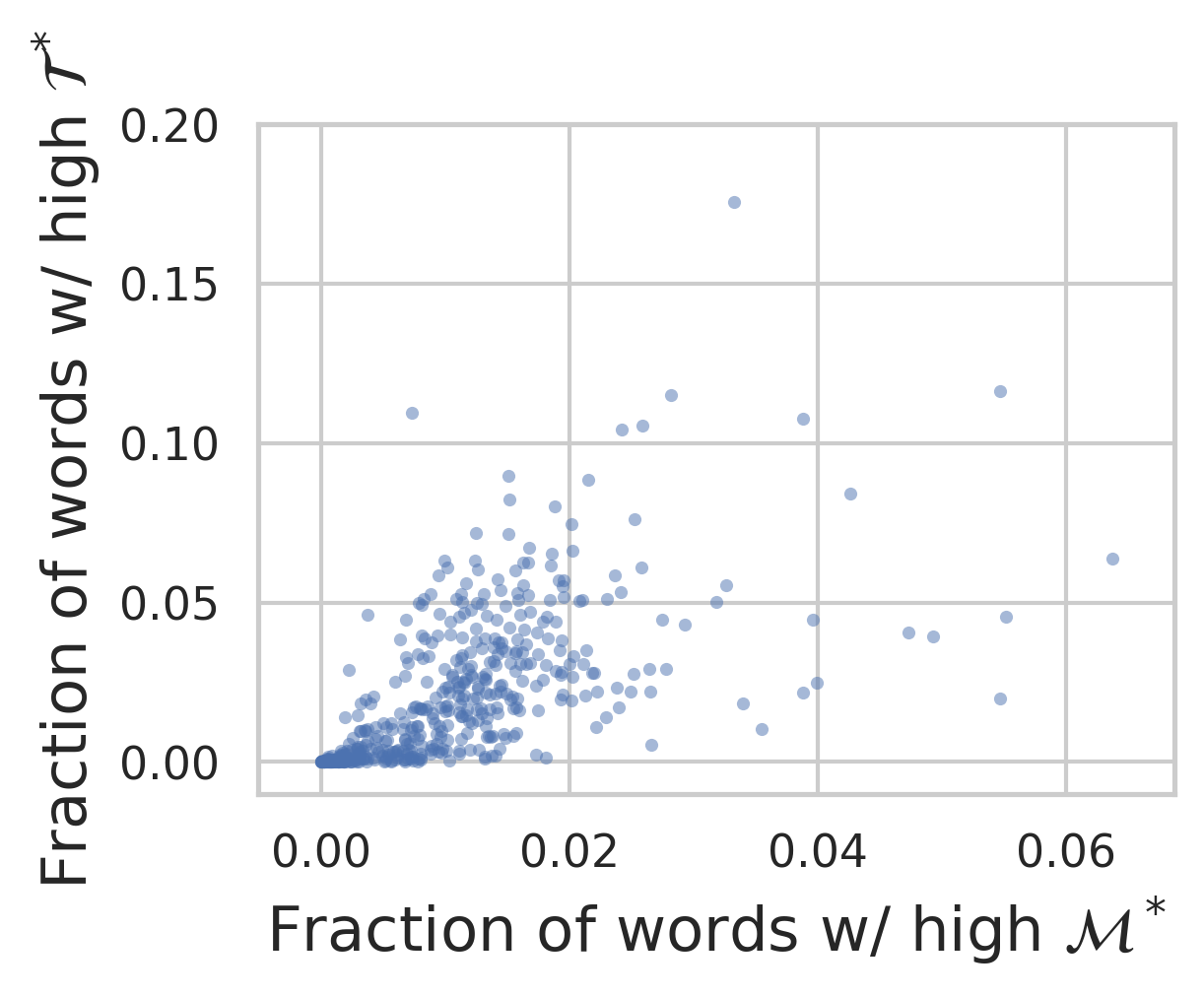}}
	\caption{A comparison of sense and type variation across subreddits, where each marker is a subreddit. The x-axis is the fraction of words with $\mathcal{M}^\ast$ in the 98th percentile, and the y-axis is the fraction of words with $\mathcal{T^\ast}$ in the 98th percentile. The subreddit r/transcribersofreddit, which had an unusually high fraction of words with high $\mathcal{T}^\ast$ (0.4101), was cropped out for visual clarity.} 
	\label{fig:sense_type}
\end{figure}

The four attributes we examine also have significant relationships with language variation when $\mathcal{F}$ is separated out into its two lexical and semantic components (the fraction of words with $\mathcal{T^\ast} > 0.3035$ and the fraction of words with $\mathcal{M}^\ast > 0.1799$). In other words, the patterns in Figure \ref{fig:users} persist when counting only unique word types and when counting only unique meanings. This is because communities with greater lexical distinctiveness also exhibit greater semantic variation (Spearman's $r_s = 0.7855$, $p < 0.001$, Figure \ref{fig:sense_type}). So, communities with strong linguistic identities express both types of variation. Further causal investigations could reveal whether the same factors, such as users' need for efficiency and expressivity, produce both unique words and unique meanings \cite{blank1999new}. 

\subsection{Topics}

\begin{figure}[t]
	\centering{\includegraphics[width=\linewidth]{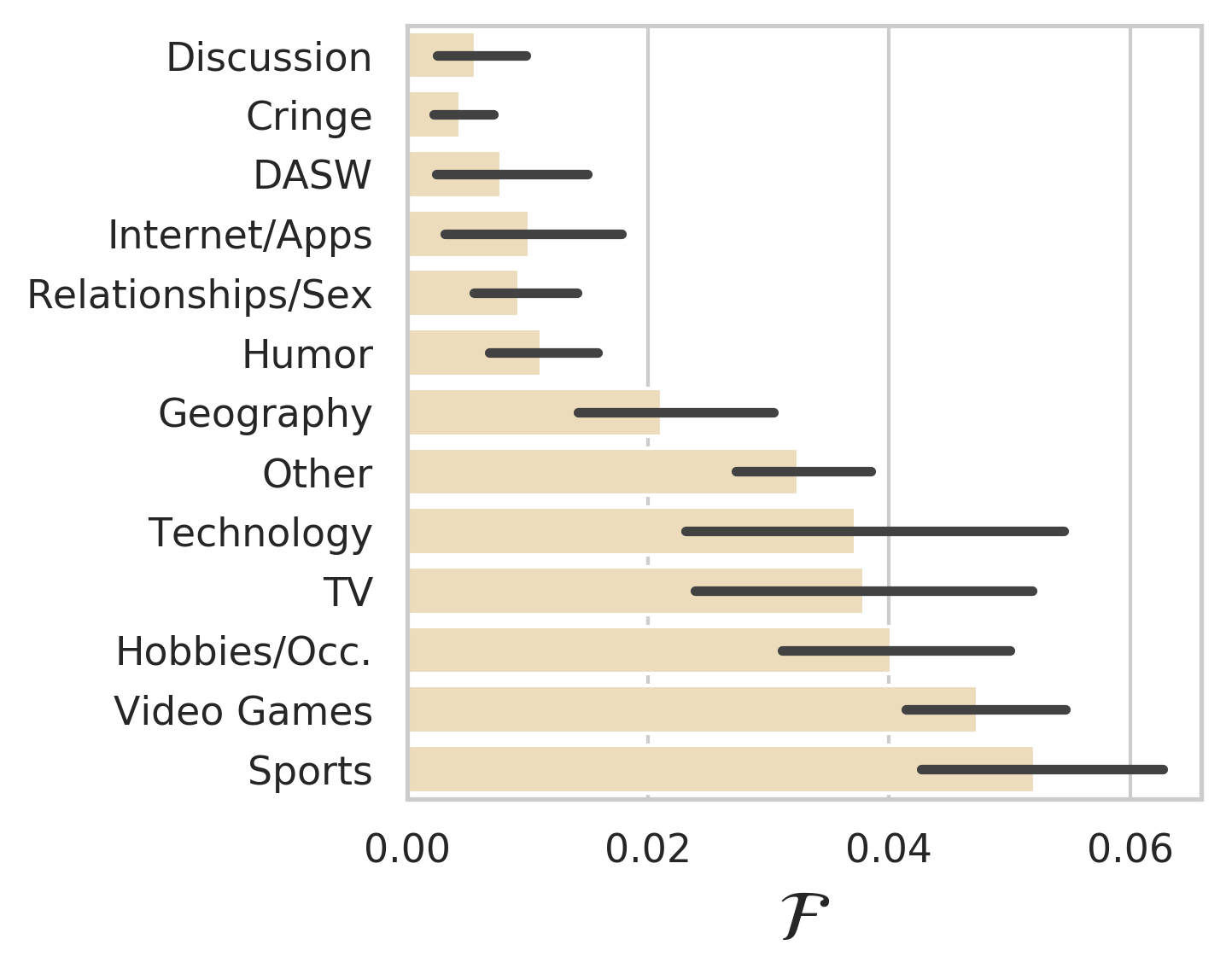}}
	\caption{A bar plot showing the average $\mathcal{F}$ of subreddits in different topics. ``DASW'' stands for the ``Disgusting/Angering/Scary/Weird'' category. Error bars are 95\% confidence intervals.} 
	\label{fig:topics}
\end{figure}

\begin{table}
\centering
\resizebox{\columnwidth}{!}{%
\begin{tabular}{p{3.5cm}cc}
\hline
& \multicolumn{2}{c}{\textit{Dependent variable:}} \\ \cline{2-3}
& \multicolumn{2}{c}{$\mathcal{F}$} \\ \hline
& (1) & (2) \\ \hline
intercept & 0.0318*** & 0.0318***\\
 & (0.001) & (0.001)\\
community size & -0.0050*** & -0.0042***\\
 & (0.001) & (0.001)\\
user activity & 0.0181*** & 0.0179***\\
 & (0.001) & (0.001)\\
user loyalty & 0.0178*** & 0.0162***\\
 & (0.001) & (0.001)\\
network density & -0.0091*** & -0.0091***\\
 & (0.001) & (0.001)\\
topic & & 0.0057***\\
 & & (0.001)\\
 \hline
Observations & 474 & 474 \\
$R^2$ & 0.505 & 0.529\\
Adjusted $R^2$ & 0.501 & 0.524\\\hline\hline
\textit{Note:} & \multicolumn{2}{c}{*$p<0.05$, **$p<0.01$, ***$p<0.001$} 
\end{tabular}
}
\caption{\label{regression} Ordinary least squares regression results for the effect of various community attributes on the fraction of community-specific words used in each community.} 
\end{table}

Language varieties can be based on interest or occupation \cite{Fishman1972, lewandowski2010sociolects}, so we also examine what topics tend to be discussed by communities with distinctive language (Figure~\ref{fig:topics}). We use r/ListofSubreddit's categorization of subreddits, focusing on the 474 subreddits in our study.\footnote{www.reddit.com/r/ListOfSubreddits/wiki/index} This categorization is hierarchical, and we choose a level of granularity so that each topic contains at least five of our subreddits. Video Games, TV, Sports, Hobbies/Occupations, and Technology tend to have more community-specific language. These communities often discuss a particular subset of the overall topic, such as a specific hobby or video game, which are rich with technical terminology. For example, r/mechanicalkeyboards ($\mathcal{F} = 0.086$) is categorized under Hobbies/Occupations. Their highly community-specific words include keyboard stores (e.g. \textit{kprepublic}), types of keyboards (e.g. \textit{ortholinear}), and keyboard components (e.g. \textit{pudding}, \textit{reds}). 

\subsection{Modeling Variation}

Finally, we run ordinary least squares regressions with attributes of Reddit communities as features and the dependent variable as communities' $\mathcal{F}$ scores. The first model has only user-based attributes as features, while the second includes a topic-related feature. These experiments help us untangle whether the topic discussed in a community has a greater impact on linguistic distinctiveness than the behaviors of the community's users. For the topic variable, we code the value as 1 if the community belongs to a topic identified as having high $\mathcal{F}$ (Technology, TV, Video Games, Hobbies/Occ., Sports, or Other), and 0 otherwise. 

Once we account for other user-based attributes, higher network density actually has a negative effect on variation (Table \ref{regression}), suggesting that its earlier marginal positive effect is due to the presence of correlated features. We find that even when a community discusses a topic that tends to have high amounts of community-specific language, attributes related to user behavior still have a bigger and more significant relationship with language use, with similar coefficients for those variables between the two models. This suggests that \textit{who} is involved in a community matters more than \textit{what} these community members discuss. 

\section{Ethical Considerations}

The Reddit posts and comments in our study are accessible by the public and were crawled by \citet{baumgartner2020pushshift}. Our project was deemed exempt from IRB review for human subjects research by the relevant administrative office at our institution. Even so, there are important ethical considerations to take when using social media data \cite{aoir2019,webb2017ethical}. Users on Reddit are not typically aware of research being conducted using their data, and therefore care needs to be taken to ensure that these users remain anonymous and unidentifable. In addition, posts and comments that are deleted by users after data collection still persist in the archived dataset. Our study minimizes risks by focusing on aggregated results, and our research questions do not involve understanding sensitive information about individual users. There is debate on whether to include direct quotes of users' content in publications \cite{webb2017ethical,vitak2016beyond}. We include a few excerpts from comments in our paper to adequately illustrate our ideas, especially since the exact wording of text can influence the predictions of NLP models, but we choose examples that do not pertain to users' personal information. 

\section{Conclusion}

We use type- and sense-based methods to detect community-specific language in Reddit communities. Our results confirm several sociolinguistic hypotheses related to the behavior of users and their use of community-specific language. Future work could develop annotated WSI datasets for online language similar to the standard \mbox{SemEval} benchmarks we used, since models developed directly on this domain may better fit its rich diversity of meanings. 

We set a foundation for further investigations on how BERT could help define unknown words or meanings in niche communities, or how linguistic norms vary across communities discussing similar topics. Our community-level analyses could be expanded to measure linguistic similarity between communities and map the dispersion of ideas among them. It is possible that the preferences of some communities towards specific senses is due to words being commonly poly\-semous and one meaning being particularly relevant to the topic of that community, while others might be linguistic innovations created by users. More research on semantic shifts may help untangle these differences. 

\section*{Acknowledgements}

We are grateful for the helpful feedback of the anonymous reviewers and our action editor, Walter Daelemans. In addition, Olivia Lewke helped us collect and organize subreddits' glossaries. This work was supported by funding from the National Science Foundation (Graduate Research Fellowship DGE-1752814 and grant IIS-1813470).

\bibliography{tacl2018}

\begin{thebibliography}{80}
\expandafter\ifx\csname natexlab\endcsname\relax\def\natexlab#1{#1}\fi

\bibitem[{Altmann et~al.(2011)Altmann, Pierrehumbert, and
  Motter}]{altmann2011niche}
Eduardo~G. Altmann, Janet~B. Pierrehumbert, and Adilson~E. Motter. 2011.
\newblock \href {https://doi.org/10.1371/journal.pone.0019009} {Niche as a
  determinant of word fate in online groups}.
\newblock \emph{PLOS One}, 6(5).

\bibitem[{Amplayo et~al.(2019)Amplayo, Hwang, and Song}]{amplayo2019autosense}
Reinald~Kim Amplayo, Seung-won Hwang, and Min Song. 2019.
\newblock \href {https://doi.org/10.1609/aaai.v33i01.33016212} {Autosense model
  for word sense induction}.
\newblock In \emph{Proceedings of the AAAI Conference on Artificial
  Intelligence}, volume~33, pages 6212--6219.

\bibitem[{Amrami and Goldberg(2018)}]{amrami2018word}
Asaf Amrami and Yoav Goldberg. 2018.
\newblock \href {https://doi.org/10.18653/v1/D18-1523} {Word sense induction
  with neural bi{LM} and symmetric patterns}.
\newblock In \emph{Proceedings of the 2018 Conference on Empirical Methods in
  Natural Language Processing}, pages 4860--4867, Brussels, Belgium.
  Association for Computational Linguistics.

\bibitem[{Amrami and Goldberg(2019)}]{amrami2019towards}
Asaf Amrami and Yoav Goldberg. 2019.
\newblock Towards better substitution-based word sense induction.
\newblock \emph{arXiv preprint arXiv:1905.12598}.

\bibitem[{Arthur and Vassilvitskii(2007)}]{10.5555/1283383.1283494}
David Arthur and Sergei Vassilvitskii. 2007.
\newblock K-means++: The advantages of careful seeding.
\newblock In \emph{Proceedings of the Eighteenth Annual ACM-SIAM Symposium on
  Discrete Algorithms}, SODA '07, page 1027–1035, USA. Society for Industrial
  and Applied Mathematics.

\bibitem[{Bamman et~al.(2014)Bamman, Dyer, and Smith}]{bamman2014distributed}
David Bamman, Chris Dyer, and Noah~A. Smith. 2014.
\newblock \href {https://doi.org/10.3115/v1/P14-2134} {Distributed
  representations of geographically situated language}.
\newblock In \emph{Proceedings of the 52nd Annual Meeting of the Association
  for Computational Linguistics (Volume 2: Short Papers)}, pages 828--834,
  Baltimore, Maryland. Association for Computational Linguistics.

\bibitem[{Baroni et~al.(2009)Baroni, Bernardini, Ferraresi, and
  Zanchetta}]{baroni2009wacky}
Marco Baroni, Silvia Bernardini, Adriano Ferraresi, and Eros Zanchetta. 2009.
\newblock \href {https://doi.org/10.1007/s10579-009-9081-4} {The wacky wide
  web: a collection of very large linguistically processed web-crawled
  corpora}.
\newblock \emph{Language Resources and Evaluation}, 43(3):209--226.

\bibitem[{Ba{\c{s}}kaya et~al.(2013)Ba{\c{s}}kaya, Sert, Cirik, and
  Yuret}]{baskaya-etal-2013-ai}
Osman Ba{\c{s}}kaya, Enis Sert, Volkan Cirik, and Deniz Yuret. 2013.
\newblock \href {https://www.aclweb.org/anthology/S13-2050} {{AI}-{KU}: Using
  substitute vectors and co-occurrence modeling for word sense induction and
  disambiguation}.
\newblock In \emph{Second Joint Conference on Lexical and Computational
  Semantics (*{SEM}), Volume 2: Proceedings of the Seventh International
  Workshop on Semantic Evaluation ({S}em{E}val 2013)}, pages 300--306, Atlanta,
  Georgia, USA. Association for Computational Linguistics.

\bibitem[{Baumgartner et~al.(2020)Baumgartner, Zannettou, Keegan, Squire, and
  Blackburn}]{baumgartner2020pushshift}
Jason Baumgartner, Savvas Zannettou, Brian Keegan, Megan Squire, and Jeremy
  Blackburn. 2020.
\newblock \href {https://www.aaai.org/ojs/index.php/ICWSM/article/view/7347}
  {The {P}ushshift {R}eddit dataset}.
\newblock In \emph{Proceedings of the International AAAI Conference on Web and
  Social Media}, volume~14, pages 830--839.

\bibitem[{Blank(1999)}]{blank1999new}
Andreas Blank. 1999.
\newblock \href {https://doi.org/10.1515/9783110804195.61} {Why do new meanings
  occur? {A} cognitive typology of the motivations for lexical semantic
  change}.
\newblock \emph{Historical Semantics and Cognition}, pages 61--90.

\bibitem[{Bouma(2009)}]{bouma2009normalized}
Gerlof Bouma. 2009.
\newblock Normalized (pointwise) mutual information in collocation extraction.
\newblock \emph{Proceedings of the German Society for Computational Linguistics
  and Language Technology (GSCL)}, pages 31--40.

\bibitem[{Brin and Page(1998)}]{brin1998anatomy}
Sergey Brin and Lawrence Page. 1998.
\newblock The anatomy of a large-scale hypertextual web search engine.
\newblock \emph{Computer Networks and ISDN Systems}, 30(1-7):107--117.

\bibitem[{Brito et~al.(1997)Brito, Chávez, Quiroz, and Yukich}]{BRITO199733}
M.R. Brito, E.L. Chávez, A.J. Quiroz, and J.E. Yukich. 1997.
\newblock \href {https://doi.org/https://doi.org/10.1016/S0167-7152(96)00213-1}
  {Connectivity of the mutual k-nearest-neighbor graph in clustering and
  outlier detection}.
\newblock \emph{Statistics \& Probability Letters}, 35(1):33 -- 42.

\bibitem[{Bucholtz and Hall(2005)}]{bucholtz2005identity}
Mary Bucholtz and Kira Hall. 2005.
\newblock \href {https://doi.org/10.1177/1461445605054407} {Identity and
  interaction: A sociocultural linguistic approach}.
\newblock \emph{Discourse Studies}, 7(4-5):585--614.

\bibitem[{Chang et~al.(2014)Chang, Pei, and Chen}]{chang-etal-2014-inducing}
Baobao Chang, Wenzhe Pei, and Miaohong Chen. 2014.
\newblock \href {https://www.aclweb.org/anthology/C14-1035} {Inducing word
  sense with automatically learned hidden concepts}.
\newblock In \emph{Proceedings of {COLING} 2014, the 25th International
  Conference on Computational Linguistics: Technical Papers}, pages 355--364,
  Dublin, Ireland. Dublin City University and Association for Computational
  Linguistics.

\bibitem[{Cohen et~al.(2014)Cohen, Delling, Pajor, and
  Werneck}]{cohen2014computing}
Edith Cohen, Daniel Delling, Thomas Pajor, and Renato~F. Werneck. 2014.
\newblock \href {https://doi.org/10.1145/2660460.2660465} {Computing classic
  closeness centrality, at scale}.
\newblock In \emph{Proceedings of the Second ACM Conference on Online Social
  Networks}, COSN '14, page 37–50, New York, NY, USA. Association for
  Computing Machinery.

\bibitem[{Danescu-Niculescu-Mizil et~al.(2013)Danescu-Niculescu-Mizil, West,
  Jurafsky, Leskovec, and Potts}]{danescu2013no}
Cristian Danescu-Niculescu-Mizil, Robert West, Dan Jurafsky, Jure Leskovec, and
  Christopher Potts. 2013.
\newblock \href {https://doi.org/10.1145/2488388.2488416} {No country for old
  members: User lifecycle and linguistic change in online communities}.
\newblock In \emph{Proceedings of the 22nd International Conference on World
  Wide Web}, WWW '13, page 307–318, New York, NY, USA. Association for
  Computing Machinery.

\bibitem[{Del~Tredici and Fern{\'a}ndez(2017)}]{del2017semantic}
Marco Del~Tredici and Raquel Fern{\'a}ndez. 2017.
\newblock \href {https://www.aclweb.org/anthology/W17-6804} {Semantic variation
  in online communities of practice}.
\newblock In \emph{{IWCS} 2017 - 12th International Conference on Computational
  Semantics - Long papers}.

\bibitem[{Del~Tredici and
  Fern{\'a}ndez(2018)}]{del-tredici-fernandez-2018-road}
Marco Del~Tredici and Raquel Fern{\'a}ndez. 2018.
\newblock \href {https://www.aclweb.org/anthology/C18-1135} {The road to
  success: Assessing the fate of linguistic innovations in online communities}.
\newblock In \emph{Proceedings of the 27th International Conference on
  Computational Linguistics}, pages 1591--1603, Santa Fe, New Mexico, USA.
  Association for Computational Linguistics.

\bibitem[{Desta(2014)}]{mashable}
Yohana Desta. 2014.
\newblock \href {https://mashable.com/2014/09/25/what-is-internet-speak/} {The
  evolution of {I}nternet speak}.
\newblock \emph{Mashable}.

\bibitem[{Devlin et~al.(2019)Devlin, Chang, Lee, and
  Toutanova}]{devlin2019bert}
Jacob Devlin, Ming-Wei Chang, Kenton Lee, and Kristina Toutanova. 2019.
\newblock \href {https://doi.org/10.18653/v1/N19-1423} {{BERT}: Pre-training of
  deep bidirectional transformers for language understanding}.
\newblock In \emph{Proceedings of the 2019 Conference of the North {A}merican
  Chapter of the Association for Computational Linguistics: Human Language
  Technologies, Volume 1 (Long and Short Papers)}, pages 4171--4186,
  Minneapolis, Minnesota. Association for Computational Linguistics.

\bibitem[{Dhuliawala et~al.(2016)Dhuliawala, Kanojia, and
  Bhattacharyya}]{dhuliawala2016slangnet}
Shehzaad Dhuliawala, Diptesh Kanojia, and Pushpak Bhattacharyya. 2016.
\newblock \href {https://www.aclweb.org/anthology/L16-1686} {{S}lang{N}et: A
  {W}ord{N}et like resource for {E}nglish slang}.
\newblock In \emph{Proceedings of the Tenth International Conference on
  Language Resources and Evaluation ({LREC}'16)}, pages 4329--4332,
  Portoro{\v{z}}, Slovenia. European Language Resources Association (ELRA).

\bibitem[{Eckert(2012)}]{eckert2012three}
Penelope Eckert. 2012.
\newblock \href {https://doi.org/10.1146/annurev-anthro-092611-145828} {Three
  waves of variation study: The emergence of meaning in the study of
  sociolinguistic variation}.
\newblock \emph{Annual Review of Anthropology}, 41(1):87--100.

\bibitem[{Eckert and McConnell-Ginet(1992)}]{eckert1992think}
Penelope Eckert and Sally McConnell-Ginet. 1992.
\newblock \href {https://doi.org/10.1146/annurev.an.21.100192.002333} {Think
  practically and look locally: Language and gender as community-based
  practice}.
\newblock \emph{Annual Review of Anthropology}, 21(1):461--488.

\bibitem[{Eisenstein(2013)}]{eisenstein-2013-bad}
Jacob Eisenstein. 2013.
\newblock \href {https://www.aclweb.org/anthology/N13-1037} {What to do about
  bad language on the internet}.
\newblock In \emph{Proceedings of the 2013 Conference of the North {A}merican
  Chapter of the Association for Computational Linguistics: Human Language
  Technologies}, pages 359--369, Atlanta, Georgia. Association for
  Computational Linguistics.

\bibitem[{Eisenstein et~al.(2014)Eisenstein, O'Connor, Smith, and
  Xing}]{eisenstein2014diffusion}
Jacob Eisenstein, Brendan O'Connor, Noah~A. Smith, and Eric~P. Xing. 2014.
\newblock \href {https://doi.org/10.1371/journal.pone.0113114} {Diffusion of
  lexical change in social media}.
\newblock \emph{PLOS ONE}, 9(11):1--13.

\bibitem[{Fishman(1972)}]{Fishman1972}
Joshua~A. Fishman. 1972.
\newblock The sociology of language.
\newblock In \emph{The Sociology of Language: An Interdisciplinary Social
  Science Approach to Language in Society}, chapter~3, pages 1--7. Newbury
  House Publishers, Rowley, MA.

\bibitem[{franzke et~al.(2020)franzke, Bechmann, Zimmer, Ess, and the
  Association~of Internet~Researchers}]{aoir2019}
aline~shakti franzke, Anja Bechmann, Michael Zimmer, Charles Ess, and the
  Association~of Internet~Researchers. 2020.
\newblock Internet research: Ethical guidelines 3.0.
\newblock \url{https://aoir.org/reports/ethics3.pdf}.

\bibitem[{Gallagher et~al.(2018)Gallagher, Reagan, Danforth, and
  Dodds}]{10.1371/journal.pone.0195644}
Ryan~J. Gallagher, Andrew~J. Reagan, Christopher~M. Danforth, and
  Peter~Sheridan Dodds. 2018.
\newblock \href {https://doi.org/10.1371/journal.pone.0195644} {Divergent
  discourse between protests and counter-protests: \#{B}lack{L}ives{M}atter and
  \#{A}ll{L}ives{M}atter}.
\newblock \emph{PLOS ONE}, 13(4):1--23.

\bibitem[{Giulianelli et~al.(2020)Giulianelli, Del~Tredici, and
  Fern{\'a}ndez}]{giulianelli-etal-2020-analysing}
Mario Giulianelli, Marco Del~Tredici, and Raquel Fern{\'a}ndez. 2020.
\newblock \href {https://doi.org/10.18653/v1/2020.acl-main.365} {Analysing
  lexical semantic change with contextualised word representations}.
\newblock In \emph{Proceedings of the 58th Annual Meeting of the Association
  for Computational Linguistics}, pages 3960--3973, Online. Association for
  Computational Linguistics.

\bibitem[{Gururangan et~al.(2020)Gururangan, Marasovi{\'c}, Swayamdipta, Lo,
  Beltagy, Downey, and Smith}]{gururangan-etal-2020-dont}
Suchin Gururangan, Ana Marasovi{\'c}, Swabha Swayamdipta, Kyle Lo, Iz~Beltagy,
  Doug Downey, and Noah~A. Smith. 2020.
\newblock \href {https://doi.org/10.18653/v1/2020.acl-main.740} {Don{'}t stop
  pretraining: Adapt language models to domains and tasks}.
\newblock In \emph{Proceedings of the 58th Annual Meeting of the Association
  for Computational Linguistics}, pages 8342--8360, Online. Association for
  Computational Linguistics.

\bibitem[{Guy(2011)}]{guy_2011}
Gregory~R. Guy. 2011.
\newblock \href {https://doi.org/10.1017/CBO9780511997068.015} {\emph{Language,
  social class, and status}}, Cambridge Handbooks in Language and Linguistics,
  chapter~10. Cambridge University Press.

\bibitem[{Hadiwinoto et~al.(2019)Hadiwinoto, Ng, and
  Gan}]{hadiwinoto-etal-2019-improved}
Christian Hadiwinoto, Hwee~Tou Ng, and Wee~Chung Gan. 2019.
\newblock \href {https://doi.org/10.18653/v1/D19-1533} {Improved word sense
  disambiguation using pre-trained contextualized word representations}.
\newblock In \emph{Proceedings of the 2019 Conference on Empirical Methods in
  Natural Language Processing and the 9th International Joint Conference on
  Natural Language Processing (EMNLP-IJCNLP)}, pages 5297--5306, Hong Kong,
  China. Association for Computational Linguistics.

\bibitem[{Hamilton et~al.(2017)Hamilton, Zhang, Danescu-Niculescu-Mizil,
  Jurafsky, and Leskovec}]{hamilton2017loyalty}
William Hamilton, Justine Zhang, Cristian Danescu-Niculescu-Mizil, Dan
  Jurafsky, and Jure Leskovec. 2017.
\newblock \href {https://aaai.org/ocs/index.php/ICWSM/ICWSM17/paper/view/15710}
  {Loyalty in online communities}.
\newblock In \emph{Proceedings of the International AAAI Conference on Web and
  Social Media}, volume~11, pages 540--543.

\bibitem[{Han and Eisenstein(2019)}]{han-eisenstein-2019-unsupervised}
Xiaochuang Han and Jacob Eisenstein. 2019.
\newblock \href {https://doi.org/10.18653/v1/D19-1433} {Unsupervised domain
  adaptation of contextualized embeddings for sequence labeling}.
\newblock In \emph{Proceedings of the 2019 Conference on Empirical Methods in
  Natural Language Processing and the 9th International Joint Conference on
  Natural Language Processing (EMNLP-IJCNLP)}, pages 4237--4247, Hong Kong,
  China. Association for Computational Linguistics.

\bibitem[{Herring and Paolillo(2006)}]{JOSL:JOSL287}
Susan~C. Herring and John~C. Paolillo. 2006.
\newblock \href {https://doi.org/10.1111/j.1467-9841.2006.00287.x} {Gender and
  genre variation in weblogs}.
\newblock \emph{Journal of Sociolinguistics}, 10(4):439--459.

\bibitem[{Holmes and Meyerhoff(1999)}]{holmes1999community}
Janet Holmes and Miriam Meyerhoff. 1999.
\newblock \href {https://doi.org/10.1017/S004740459900202X} {The community of
  practice: Theories and methodologies in language and gender research}.
\newblock \emph{Language in Society}, 28(2):173–183.

\bibitem[{Hu et~al.(2019)Hu, Li, and Liang}]{hu2019diachronic}
Renfen Hu, Shen Li, and Shichen Liang. 2019.
\newblock \href {https://doi.org/10.18653/v1/P19-1379} {Diachronic sense
  modeling with deep contextualized word embeddings: An ecological view}.
\newblock In \emph{Proceedings of the 57th Annual Meeting of the Association
  for Computational Linguistics}, pages 3899--3908, Florence, Italy.
  Association for Computational Linguistics.

\bibitem[{{Jianbo Shi} and {Malik}(2000)}]{spectral2000}
{Jianbo Shi} and J.~{Malik}. 2000.
\newblock \href {https://doi.org/10.1109/34.868688} {Normalized cuts and image
  segmentation}.
\newblock \emph{IEEE Transactions on Pattern Analysis and Machine
  Intelligence}, 22(8):888--905.

\bibitem[{Jones et~al.(2004)Jones, Ravid, and Rafaeli}]{jones2004information}
Quentin Jones, Gilad Ravid, and Sheizaf Rafaeli. 2004.
\newblock \href {https://doi.org/10.1287/isre.1040.0023} {Information overload
  and the message dynamics of online interaction spaces: A theoretical model
  and empirical exploration}.
\newblock \emph{Information Systems Research}, 15(2):194--210.

\bibitem[{Jurgens and Klapaftis(2013)}]{jurgens-klapaftis-2013-semeval}
David Jurgens and Ioannis Klapaftis. 2013.
\newblock \href {https://www.aclweb.org/anthology/S13-2049} {{S}em{E}val-2013
  task 13: Word sense induction for graded and non-graded senses}.
\newblock In \emph{Second Joint Conference on Lexical and Computational
  Semantics (*{SEM}), Volume 2: Proceedings of the Seventh International
  Workshop on Semantic Evaluation ({S}em{E}val 2013)}, pages 290--299, Atlanta,
  Georgia, USA. Association for Computational Linguistics.

\bibitem[{Labov(1973)}]{labov1973linguistic}
William Labov. 1973.
\newblock \href {http://www.jstor.org/stable/4166708} {The linguistic
  consequences of being a lame}.
\newblock \emph{Language in Society}, 2(1):81--115.

\bibitem[{Lau et~al.(2013)Lau, Cook, and Baldwin}]{lau2013unimelb}
Jey~Han Lau, Paul Cook, and Timothy Baldwin. 2013.
\newblock \href {https://www.aclweb.org/anthology/S13-2051} {unimelb: Topic
  modelling-based word sense induction}.
\newblock In \emph{Second Joint Conference on Lexical and Computational
  Semantics (*{SEM}), Volume 2: Proceedings of the Seventh International
  Workshop on Semantic Evaluation ({S}em{E}val 2013)}, pages 307--311, Atlanta,
  Georgia, USA. Association for Computational Linguistics.

\bibitem[{Leavitt(2015)}]{10.1145/2675133.2675175}
Alex Leavitt. 2015.
\newblock \href {https://doi.org/10.1145/2675133.2675175} {"this is a throwaway
  account": Temporary technical identities and perceptions of anonymity in a
  massive online community}.
\newblock In \emph{Proceedings of the 18th ACM Conference on Computer Supported
  Cooperative Work \& Social Computing}, CSCW '15, page 317–327, New York,
  NY, USA. Association for Computing Machinery.

\bibitem[{Lewandowski(2010)}]{lewandowski2010sociolects}
Marcin Lewandowski. 2010.
\newblock \href {https://doi.org/https://doi.org/10.14746/il.2010.20.6}
  {Sociolects and registers--a contrastive analysis of two kinds of linguistic
  variation}.
\newblock \emph{Investigationes Linguisticae}, 20:60--79.

\bibitem[{{Lin}(1991)}]{Lin1991jsd}
J.~{Lin}. 1991.
\newblock \href {https://doi.org/10.1109/18.61115} {Divergence measures based
  on the shannon entropy}.
\newblock \emph{IEEE Transactions on Information Theory}, 37(1):145--151.

\bibitem[{{Lloyd}(1982)}]{kmeans1982}
S.~{Lloyd}. 1982.
\newblock \href {https://doi.org/10.1109/TIT.1982.1056489} {Least squares
  quantization in pcm}.
\newblock \emph{IEEE Transactions on Information Theory}, 28(2):129--137.

\bibitem[{Lu et~al.(2020)Lu, Henchion, and Mac~Namee}]{lu-etal-2020-diverging}
Jinghui Lu, Maeve Henchion, and Brian Mac~Namee. 2020.
\newblock \href {https://www.aclweb.org/anthology/2020.lrec-1.832} {Diverging
  divergences: Examining variants of {J}ensen {S}hannon divergence for corpus
  comparison tasks}.
\newblock In \emph{Proceedings of the 12th Language Resources and Evaluation
  Conference}, pages 6740--6744, Marseille, France. European Language Resources
  Association.

\bibitem[{Lui and Baldwin(2012)}]{lui-baldwin-2012-langid}
Marco Lui and Timothy Baldwin. 2012.
\newblock \href {https://www.aclweb.org/anthology/P12-3005} {langid.py: An
  off-the-shelf language identification tool}.
\newblock In \emph{Proceedings of the {ACL} 2012 System Demonstrations}, pages
  25--30, Jeju Island, Korea. Association for Computational Linguistics.

\bibitem[{von Luxburg(2007)}]{VonLuxburg2007}
Ulrike von Luxburg. 2007.
\newblock \href {https://doi.org/10.1007/s11222-007-9033-z} {{A tutorial on
  spectral clustering}}.
\newblock \emph{Statistics and Computing}, 17(4):395--416.

\bibitem[{Magalhães(2019)}]{unbabel}
Raquel Magalhães. 2019.
\newblock \href {https://unbabel.com/blog/speak-internet-slang/} {Do you speak
  internet? {H}ow internet slang is changing language}.
\newblock \emph{Understanding with Unbabel}.

\bibitem[{Manandhar and Klapaftis(2009)}]{manandhar-klapaftis-2009-semeval}
Suresh Manandhar and Ioannis Klapaftis. 2009.
\newblock \href {https://www.aclweb.org/anthology/W09-2419} {{S}em{E}val-2010
  task 14: Evaluation setting for word sense induction {\&} disambiguation
  systems}.
\newblock In \emph{Proceedings of the Workshop on Semantic Evaluations: Recent
  Achievements and Future Directions ({SEW}-2009)}, pages 117--122, Boulder,
  Colorado. Association for Computational Linguistics.

\bibitem[{Manning et~al.(2008)Manning, Raghavan, and
  Sch{\"u}tze}]{manning2008introduction}
Christopher~D. Manning, Prabhakar Raghavan, and Hinrich Sch{\"u}tze. 2008.
\newblock \emph{Introduction to Information Retrieval}.
\newblock Cambridge University Press.

\bibitem[{Mihalcea and Tarau(2004)}]{mihalcea-tarau-2004-textrank}
Rada Mihalcea and Paul Tarau. 2004.
\newblock \href {https://www.aclweb.org/anthology/W04-3252} {{T}ext{R}ank:
  Bringing order into text}.
\newblock In \emph{Proceedings of the 2004 Conference on Empirical Methods in
  Natural Language Processing}, pages 404--411, Barcelona, Spain. Association
  for Computational Linguistics.

\bibitem[{Miller et~al.(1993)Miller, Leacock, Tengi, and
  Bunker}]{miller1993semantic}
George~A. Miller, Claudia Leacock, Randee Tengi, and Ross~T. Bunker. 1993.
\newblock \href {https://www.aclweb.org/anthology/H93-1061} {A semantic
  concordance}.
\newblock In \emph{{H}uman {L}anguage {T}echnology: Proceedings of a Workshop},
  Plainsboro, New Jersey.

\bibitem[{Milroy(1987)}]{milroy1987language}
L.~Milroy. 1987.
\newblock \emph{Language and Social Networks}.
\newblock Language in Society. Wiley-Blackwell, Oxford.

\bibitem[{Milroy and Milroy(1992)}]{milroy1992social}
Lesley Milroy and James Milroy. 1992.
\newblock \href {http://www.jstor.org/stable/4168309} {Social network and
  social class: Toward an integrated sociolinguistic model}.
\newblock \emph{Language in Society}, 21(1):1--26.

\bibitem[{Nguyen et~al.(2016)Nguyen, Do{\u{g}}ru{\"o}z, Ros{\'e}, and
  de~Jong}]{nguyen2016computational}
Dong Nguyen, A.~Seza Do{\u{g}}ru{\"o}z, Carolyn~P. Ros{\'e}, and Franciska
  de~Jong. 2016.
\newblock \href {https://doi.org/10.1162/COLI_a_00258} {Computational
  sociolinguistics: A {S}urvey}.
\newblock \emph{Computational Linguistics}, 42(3):537--593.

\bibitem[{Nguyen and Ros{\'e}(2011)}]{nguyen2011language}
Dong Nguyen and Carolyn~P. Ros{\'e}. 2011.
\newblock \href {https://www.aclweb.org/anthology/W11-0710} {Language use as a
  reflection of socialization in online communities}.
\newblock In \emph{Proceedings of the Workshop on Language in Social Media
  ({LSM} 2011)}, pages 76--85, Portland, Oregon. Association for Computational
  Linguistics.

\bibitem[{Pechenick et~al.(2015)Pechenick, Danforth, and
  Dodds}]{10.1371/journal.pone.0137041}
Eitan~Adam Pechenick, Christopher~M. Danforth, and Peter~Sheridan Dodds. 2015.
\newblock \href {https://doi.org/10.1371/journal.pone.0137041} {Characterizing
  the google books corpus: Strong limits to inferences of socio-cultural and
  linguistic evolution}.
\newblock \emph{PLOS ONE}, 10(10):1--24.

\bibitem[{Pedregosa et~al.(2011)Pedregosa, Varoquaux, Gramfort, Michel,
  Thirion, Grisel, Blondel, Prettenhofer, Weiss, Dubourg, Vanderplas, Passos,
  Cournapeau, Brucher, Perrot, and Duchesnay}]{scikit-learn}
F.~Pedregosa, G.~Varoquaux, A.~Gramfort, V.~Michel, B.~Thirion, O.~Grisel,
  M.~Blondel, P.~Prettenhofer, R.~Weiss, V.~Dubourg, J.~Vanderplas, A.~Passos,
  D.~Cournapeau, M.~Brucher, M.~Perrot, and E.~Duchesnay. 2011.
\newblock Scikit-learn: Machine learning in {P}ython.
\newblock \emph{Journal of Machine Learning Research}, 12:2825--2830.

\bibitem[{Pei et~al.(2019)Pei, Sun, and Xu}]{pei2019slang}
Zhengqi Pei, Zhewei Sun, and Yang Xu. 2019.
\newblock \href {https://doi.org/10.18653/v1/K19-1082} {Slang detection and
  identification}.
\newblock In \emph{Proceedings of the 23rd Conference on Computational Natural
  Language Learning (CoNLL)}, pages 881--889, Hong Kong, China. Association for
  Computational Linguistics.

\bibitem[{Pelleg and Moore(2000)}]{pelleg2000x}
Dan Pelleg and Andrew~W. Moore. 2000.
\newblock X-means: Extending k-means with efficient estimation of the number of
  clusters.
\newblock In \emph{Proceedings of the Seventeenth International Conference on
  Machine Learning}, ICML '00, page 727–734, San Francisco, CA, USA. Morgan
  Kaufmann Publishers Inc.

\bibitem[{Postmes et~al.(2000)Postmes, Spears, and Lea}]{postmes2000formation}
Tom Postmes, Russell Spears, and Martin Lea. 2000.
\newblock \href {https://doi.org/10.1111/j.1468-2958.2000.tb00761.x} {The
  formation of group norms in computer-mediated communication}.
\newblock \emph{Human Communication Research}, 26(3):341--371.

\bibitem[{Reif et~al.(2019)Reif, Yuan, Wattenberg, Viegas, Coenen, Pearce, and
  Kim}]{NIPS2019_9065}
Emily Reif, Ann Yuan, Martin Wattenberg, Fernanda~B. Viegas, Andy Coenen, Adam
  Pearce, and Been Kim. 2019.
\newblock \href
  {http://papers.nips.cc/paper/9065-visualizing-and-measuring-the-geometry-of-bert.pdf}
  {Visualizing and measuring the geometry of {BERT}}.
\newblock In \emph{Advances in Neural Information Processing Systems 32}, pages
  8594--8603.

\bibitem[{Rotabi and Kleinberg(2016)}]{rotabi2016status}
Rahmtin Rotabi and Jon Kleinberg. 2016.
\newblock \href
  {https://www.aaai.org/ocs/index.php/ICWSM/ICWSM16/paper/view/13053} {The
  status gradient of trends in social media}.
\newblock In \emph{Proceedings of the International AAAI Conference on Web and
  Social Media}, volume~10, pages 319--328.

\bibitem[{Sharma and Dodsworth(2020)}]{sharma2020language}
Devyani Sharma and Robin Dodsworth. 2020.
\newblock \href {https://doi.org/10.1146/annurev-linguistics-011619-030524}
  {Language variation and social networks}.
\newblock \emph{Annual Review of Linguistics}, 6(1):341--361.

\bibitem[{Song et~al.(2016)Song, Wang, Mi, and Gildea}]{song2016sense}
Linfeng Song, Zhiguo Wang, Haitao Mi, and Daniel Gildea. 2016.
\newblock \href {https://doi.org/10.18653/v1/S16-2009} {Sense embedding
  learning for word sense induction}.
\newblock In \emph{Proceedings of the Fifth Joint Conference on Lexical and
  Computational Semantics}, pages 85--90, Berlin, Germany. Association for
  Computational Linguistics.

\bibitem[{Stewart et~al.(2017)Stewart, Chancellor, De~Choudhury, and
  Eisenstein}]{stewart2017anorexia}
Ian Stewart, Stevie Chancellor, Munmun De~Choudhury, and Jacob Eisenstein.
  2017.
\newblock \href {https://doi.org/10.1109/BigData.2017.8258465} {\#anorexia,
  \#anarexia, \#anarexyia: Characterizing online community practices with
  orthographic variation}.
\newblock In \emph{2017 IEEE International Conference on Big Data (Big Data)},
  pages 4353--4361.

\bibitem[{Stewart and Eisenstein(2018)}]{stewart2018making}
Ian Stewart and Jacob Eisenstein. 2018.
\newblock \href {https://doi.org/10.18653/v1/D18-1467} {Making {``}fetch{''}
  happen: The influence of social and linguistic context on nonstandard word
  growth and decline}.
\newblock In \emph{Proceedings of the 2018 Conference on Empirical Methods in
  Natural Language Processing}, pages 4360--4370, Brussels, Belgium.
  Association for Computational Linguistics.

\bibitem[{Sun et~al.(2014)Sun, Rau, and Ma}]{sun2014understanding}
Na~Sun, Patrick Pei-Luen Rau, and Liang Ma. 2014.
\newblock \href {https://doi.org/10.1016/j.chb.2014.05.022} {Understanding
  lurkers in online communities: A literature review}.
\newblock \emph{Comput. Hum. Behav.}, 38:110–117.

\bibitem[{Tibshirani et~al.(2001)Tibshirani, Walther, and
  Hastie}]{tibshirani2001estimating}
Robert Tibshirani, Guenther Walther, and Trevor Hastie. 2001.
\newblock \href {https://doi.org/10.1111/1467-9868.00293} {Estimating the
  number of clusters in a data set via the gap statistic}.
\newblock \emph{Journal of the Royal Statistical Society: Series B (Statistical
  Methodology)}, 63(2):411--423.

\bibitem[{Vitak et~al.(2016)Vitak, Shilton, and Ashktorab}]{vitak2016beyond}
Jessica Vitak, Katie Shilton, and Zahra Ashktorab. 2016.
\newblock \href {https://doi.org/10.1145/2818048.2820078} {Beyond the belmont
  principles: Ethical challenges, practices, and beliefs in the online data
  research community}.
\newblock In \emph{Proceedings of the 19th ACM Conference on Computer-Supported
  Cooperative Work \& Social Computing}, pages 941--953.

\bibitem[{Voorhees(1999)}]{voorhees1999trec}
Ellen~M. Voorhees. 1999.
\newblock \href {https://trec.nist.gov/pubs/trec8/papers/qa_report.pdf} {The
  {TREC}-8 question answering track report}.
\newblock In \emph{Proceedings of the 8th Text Retrieval Conference (TREC-8)}.

\bibitem[{Webb et~al.(2017)Webb, Jirotka, Stahl, Housley, Edwards, Williams,
  Procter, Rana, and Burnap}]{webb2017ethical}
Helena Webb, Marina Jirotka, Bernd~Carsten Stahl, William Housley, Adam
  Edwards, Matthew Williams, Rob Procter, Omer Rana, and Pete Burnap. 2017.
\newblock \href {https://doi.org/10.1145/3091478.3091489} {The ethical
  challenges of publishing {T}witter data for research dissemination}.
\newblock In \emph{Proceedings of the 2017 ACM on Web Science Conference},
  WebSci '17, page 339–348, New York, NY, USA. Association for Computing
  Machinery.

\bibitem[{Wenger(2000)}]{wenger1998communities}
Etienne Wenger. 2000.
\newblock \href {https://doi.org/10.1177/135050840072002} {Communities of
  practice and social learning systems}.
\newblock \emph{Organization}, 7(2):225--246.

\bibitem[{Wiedemann et~al.(2019)Wiedemann, Remus, Chawla, and
  Biemann}]{wiedemann-etal-2019-does}
Gregor Wiedemann, Steffen Remus, Avi Chawla, and Chris Biemann. 2019.
\newblock \href
  {https://corpora.linguistik.uni-erlangen.de/data/konvens/proceedings/papers/KONVENS2019_paper_43.pdf}
  {Does {BERT} make any sense? {I}nterpretable word sense disambiguation with
  contextualized embeddings}.
\newblock In \emph{Proceedings of the 15th Conference on Natural Language
  Processing (KONVENS 2019): Long Papers}, pages 161--170, Erlangen, Germany.
  German Society for Computational Linguistics \& Language Technology.

\bibitem[{Wolf et~al.(2020)Wolf, Debut, Sanh, Chaumond, Delangue, Moi, Cistac,
  Rault, Louf, Funtowicz, Davison, Shleifer, von Platen, Ma, Jernite, Plu, Xu,
  Le~Scao, Gugger, Drame, Lhoest, and Rush}]{wolf-etal-2020-transformers}
Thomas Wolf, Lysandre Debut, Victor Sanh, Julien Chaumond, Clement Delangue,
  Anthony Moi, Pierric Cistac, Tim Rault, Remi Louf, Morgan Funtowicz, Joe
  Davison, Sam Shleifer, Patrick von Platen, Clara Ma, Yacine Jernite, Julien
  Plu, Canwen Xu, Teven Le~Scao, Sylvain Gugger, Mariama Drame, Quentin Lhoest,
  and Alexander Rush. 2020.
\newblock \href {https://doi.org/10.18653/v1/2020.emnlp-demos.6} {Transformers:
  State-of-the-art natural language processing}.
\newblock In \emph{Proceedings of the 2020 Conference on Empirical Methods in
  Natural Language Processing: System Demonstrations}, pages 38--45, Online.
  Association for Computational Linguistics.

\bibitem[{Yang and Eisenstein(2017)}]{yang2017overcoming}
Yi~Yang and Jacob Eisenstein. 2017.
\newblock \href {https://doi.org/10.1162/tacl_a_00062} {Overcoming language
  variation in sentiment analysis with social attention}.
\newblock \emph{Transactions of the Association for Computational Linguistics},
  5:295--307.

\bibitem[{Zhang et~al.(2017)Zhang, Hamilton, Danescu-Niculescu-Mizil, Jurafsky,
  and Leskovec}]{zhang2017community}
Justine Zhang, William~L. Hamilton, Cristian Danescu-Niculescu-Mizil, Dan
  Jurafsky, and Jure Leskovec. 2017.
\newblock \href {https://aaai.org/ocs/index.php/ICWSM/ICWSM17/paper/view/15706}
  {Community identity and user engagement in a multi-community landscape}.
\newblock In \emph{Proceedings of the International AAAI Conference on Web and
  Social Media}, volume~11, pages 377--386.

\end{thebibliography}
\bibliographystyle{acl_natbib}

\end{document}